\documentclass[11pt]{article}

\usepackage[utf8]{inputenc}
\usepackage[T1]{fontenc}
\usepackage{microtype}
\usepackage{nicefrac}
\usepackage{placeins}
\usepackage{algorithm}
\usepackage{algpseudocode}
\usepackage{graphicx,subcaption} 
\usepackage{amsfonts,amssymb,amsmath,amsthm,amsopn}	
\usepackage{hhline,booktabs,colortbl,multirow,tabularx,threeparttable} 
\usepackage[numbers,compress]{natbib}

\usepackage{enumitem}
\usepackage{float}
\usepackage{hyperref}
\usepackage{bookmark}
\usepackage{prettyref}
\usepackage{setspace}
\usepackage{color}
\usepackage{xcolor}  
\usepackage{titlesec} 
\usepackage{reporttitleblock} 
\usepackage{geometry}

\usepackage{tikz}
\usepackage{pgfplots}
\usetikzlibrary{positioning,shapes,shadows,arrows,arrows.meta,calc,fit,backgrounds,decorations.pathmorphing}
\tikzstyle{component}=[rectangle, draw=black, rounded corners, fill=blue!40, drop shadow, text centered, anchor=north, text=white, minimum height=1cm]
\tikzstyle{arrow}=[->, thick]

\pgfplotsset{compat=1.18}
\usetikzlibrary{intersections}
\usetikzlibrary{pgfplots.statistics}
\usepgfplotslibrary{fillbetween}

\usepackage{eso-pic}   
\usepackage{titling}   
\usepackage{fancyhdr}  
\usepackage{lastpage}  
\usepackage{fontawesome5} 

\newcommand{\PageCounterText}{%
  {\small\color{headerblue}\faBookOpen~~\textbf{\thepage}/\pageref{LastPage}}%
}
\newcommand{\FirstPageFooterLeft}{%
  \small{\color{headerblue}\faIcon{book}}~~%
  {\small\color{headerblue}\textbf{\ReportTitleMarkText\ | \today}}%
}
\fancypagestyle{allpagesfooter}{%
  \fancyhf{}%
  \fancyfoot[R]{\PageCounterText}%
}
\fancypagestyle{firstpagefooter}{%
  \fancyhf{}%
  \fancyfoot[L]{\FirstPageFooterLeft}%
  \fancyfoot[R]{\PageCounterText}%
}
\definecolor{myblue}{RGB}{34,31,217}
\definecolor{mycyan}{gray}{.7}
\definecolor{Gray}{gray}{0.9}
\definecolor{headerblue}{RGB}{0,85,150}    
\definecolor{natureblue}{RGB}{11,61,145}   
\definecolor{titlegray}{RGB}{80,80,80}     
\definecolor{titlebox}{RGB}{245,248,252}   
\newlength{\abstractpad}
\hypersetup{
  colorlinks=true,
  linkcolor=natureblue,
  citecolor=natureblue,
  urlcolor=natureblue,
  filecolor=natureblue
}

\newcommand{\our}{\texttt{REFLEX}}

\let\cite\citep
\definecolor{cvaccent}{HTML}{A23B2A}
\definecolor{cvsubtle}{HTML}{243B5A}
\colorlet{cvaccentsoft}{cvaccent!10}
\colorlet{cvsubtlesoft}{cvsubtle!10}
\colorlet{cvline}{cvaccent!55}
\colorlet{cvneutral}{black!65}

\ReportClearAffils{}
\ReportClearAuthors{}
\ReportTitle{REFLEX: RNA Ensemble Generation via Flexibility-Calibrated Stochastic Bridge}
\ReportTitleMark{COLALab Report}
\ReportAcceptedNote{}
\ReportAffil{1}{Earlham Institute, United Kingdom.}
\ReportAffil{2}{Living Systems Institute, University of Exeter, United Kingdom.}
\ReportAffil{3}{Department of Computer Science, University of Exeter, United Kingdom.}
\ReportAffil{4}{School of Computing Sciences, University of East Anglia, United Kingdom.}
\ReportAuthor{Wanli Ma}{1,2,3}{wm385@exeter.ac.uk}
\ReportAuthor{Tianmeng Hu}{1,2,3}{tianmeng0824@gmail.com}
\ReportAuthor{Ke Li}{1,2,3,4}[corresponding]{Ke.Li@earlham.ac.uk}

\titleformat{\section}{\color{natureblue}\normalfont\Large\bfseries}{\thesection}{1em}{}
\titleformat{\subsection}{\color{natureblue}\normalfont\large\bfseries}{\thesubsection}{1em}{}
\titleformat{\subsubsection}{\color{natureblue}\normalfont\normalsize\bfseries}{\thesubsubsection}{1em}{}

\newcommand{\pref}{\prettyref}

\newrefformat{fig}{Figure~\ref{#1}}
\newrefformat{tab}{Table~\ref{#1}}
\newrefformat{sec}{Section~\ref{#1}}
\newrefformat{app}{Appendix~\ref{#1}}
\newrefformat{alg}{Algorithm~\ref{#1}}
\newrefformat{property}{Property~\ref{#1}}
\newrefformat{theorem}{Theorem~\ref{#1}}
\newrefformat{corollary}{Corollary~\ref{#1}}
\newrefformat{proposition}{Proposition~\ref{#1}}
\newrefformat{def}{Definition~\ref{#1}}
\newrefformat{eq}{Eq.~(\ref{#1})}
\newrefformat{ass}{Assumption~\ref{#1}}

\usepackage{lscape}

\begin{document}
\thispagestyle{firstpagefooter}

\AddToShipoutPictureBG*{%
  \AtPageUpperLeft{%
    \raisebox{-2.6cm}{\hspace{2cm}%
      \parbox[t]{\textwidth}{%
        \includegraphics[height=1.4cm]{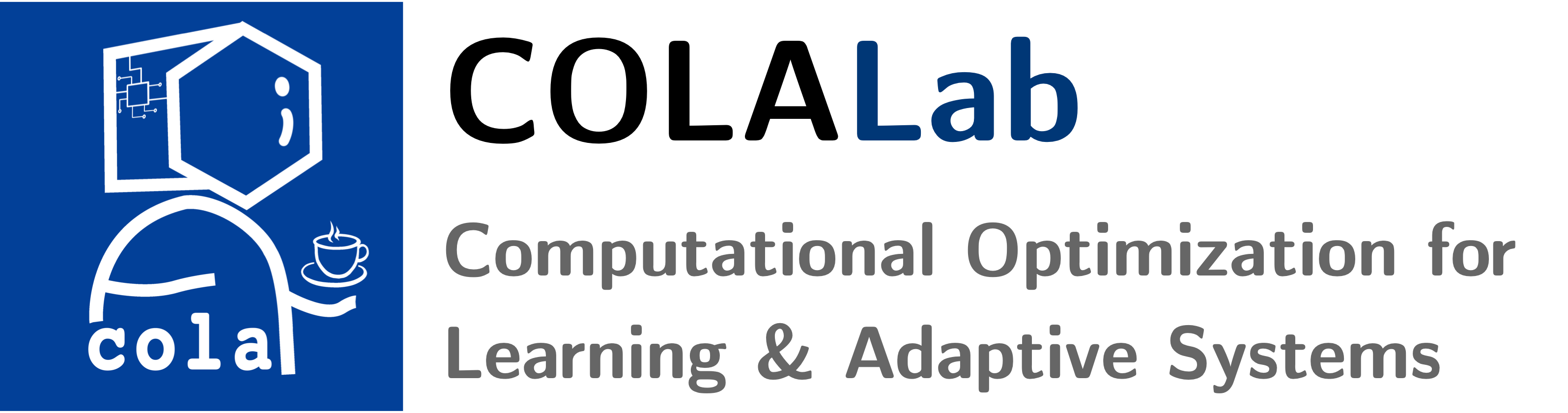}
      }%
    }%
  }%
}

\vspace*{0.8cm}

\MakeReportTitleBlock{}

\vspace{1.0em}

\noindent
	\begin{tikzpicture}
	  \node[
	    fill=titlebox,
	    rounded corners=10pt,
	    inner xsep=\abstractpad,
	    inner ysep=10pt
	  ]{%
	    \begin{minipage}{\dimexpr\textwidth-2\abstractpad\relax}
	      \setlength{\parindent}{0pt}%
	      {\centering\large\color{natureblue}\textbf{Abstract}\par}
\vspace{0.4em}
\noindent
RNA molecules often function through multiple experimentally observed conformational states. A central challenge is to capture ensemble diversity without sacrificing structural fidelity: samples may collapse toward averaged representative structures or drift into unsupported conformational regions. We introduce \our\ (\underline{R}NA \underline{E}nsemble generation calibrated by \underline{FLEX}ibility), a sequence-conditioned generator for RNA backbone-frame ensembles based on a heteroscedastic stochastic bridge on \(\mathrm{SE}(3)\). During training, experimental \(\mathrm{B}\)-factors calibrate residue-wise bridge widths, exposing each observed conformer through local perturbation neighborhoods with residue-specific scales while keeping constrained regions sharper. A learned flexibility condition guides the bridge-field network to recover target conformer-specific frames from these calibrated noisy states. At inference, \our\ uses only the input sequence, requiring no MSA-subsampling heuristics. On a multi-conformer RNA benchmark, \our\ achieves the best precision--recall trade-off for ensemble coverage across evaluation thresholds, while recovery-stress ablations indicate gains in target-frame recovery under stronger perturbations.
\par

	    \end{minipage}
	  };
	\end{tikzpicture}



\section{Introduction}
\label{sec:introduction}

RNA molecules are intrinsically dynamic, often interconverting among multiple conformational states. Such ensembles are central to many RNA functions, including ligand recognition, molecular switching, and regulatory control~\cite{mustoe2014hierarchy,ganser2019roles,maruvsivc2019rna,ken2023rna}. Sequence-conditioned RNA modeling should therefore move beyond predicting one representative structure and aim to recover multiple plausible conformational states while preserving the local constraints that make each state structurally valid.

Recent deep learning methods have substantially advanced RNA $\mathrm{3D}$ modeling. \texttt{\textsc{AlphaFold3}}~\cite{AbramsonADEGPRWBB24} and \texttt{\textsc{RhoFold+}}~\cite{ShenHSLWWCWHXZKKWYCL24} achieve strong accuracy when a single representative conformation is sufficient. Alternative conformations can sometimes be elicited by repeated inference under perturbed evolutionary context, such as MSA subsampling~\cite{SteeleOOAPHOCK24}, but these procedures rely on inference-time search to explore conformational alternatives. Dedicated generative methods have begun to model RNA ensembles more directly: \texttt{\textsc{RNAbpFlow}}~\cite{TarafderB25} generates RNA backbones conditioned on sequence and predicted base-pair information, while \texttt{\textsc{DynaRNA}}~\cite{li2025dynarna} models conformational ensembles through coordinate-space diffusion from structural inputs. Yet existing approaches do not explicitly calibrate how conformational uncertainty should vary across nucleotides.

This local calibration matters because RNA flexibility is spatially heterogeneous. Base-paired helices are often constrained, whereas bulges, loops, junctions, and hinge regions can undergo larger-amplitude motion~\cite{carugo2018atomic,guruge2018b,maruvsivc2019rna}. Thus, the same sequence can have both rigid regions with narrow coordinate support and dynamic regions with multiple plausible conformational targets. Without a training signal for this local ambiguity, stochastic generators face a precision--recall tension: they may collapse toward dominant conformations or spread samples into unsupported structures. A sequence-to-ensemble generator should therefore learn where structural uncertainty is expected and allocate it as a nucleotide-wise property of the molecule.

\begin{figure}[!t]
\centering
\includegraphics[width=\linewidth]{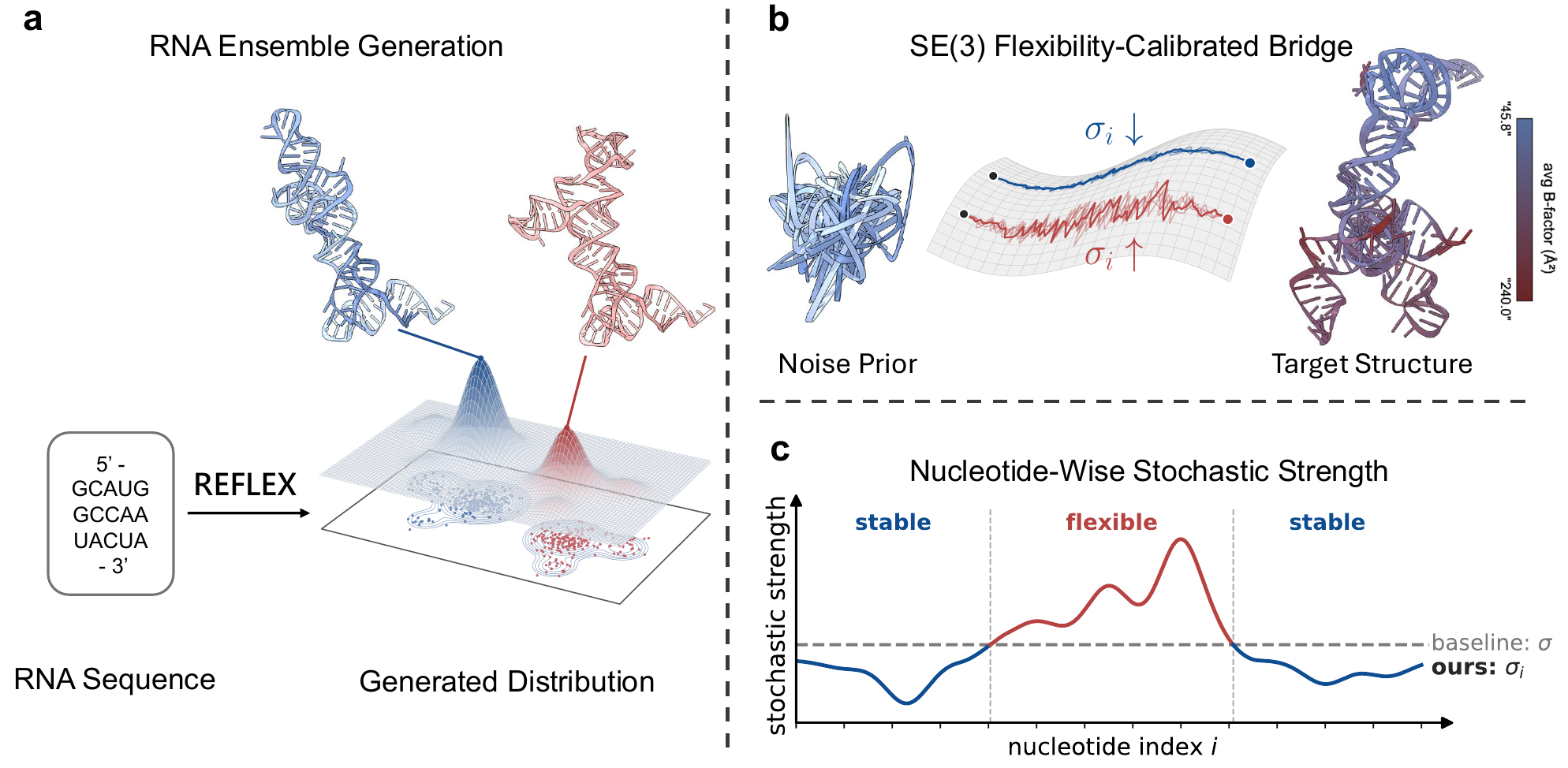}
\caption{\textbf{Overview of \our.} \textbf{a}: RNA ensemble generation aims to recover accurate and diverse conformational alternatives. \textbf{b}: Flexibility-calibrated stochastic bridge on \(\mathrm{SE}(3)\). The structures on the right are multiple target conformations, and their averaged \(\mathrm{B}\)-factor profile captures nucleotide-wise flexibility used to calibrate the stochastic bridge. \textbf{c}: Nucleotide-wise stochastic strength, with nucleotides on the horizontal axis and stochasticity strength on the vertical axis; flexible positions receive stronger stochasticity than constrained positions.}
\label{fig:framework}
\end{figure}

We introduce~\our, a sequence-conditioned stochastic bridge on $\mathrm{SE}(3)$ for RNA backbone-frame ensemble generation. During training, \our\ uses chain-normalized experimental $\mathrm{B}$-factors as a proxy for relative local mobility and converts them into nucleotide-wise bridge widths. Flexible nucleotides receive broader bridge perturbations, while rigid regions receive narrower ones. At inference, a sequence-derived latent flexibility condition guides the generative dynamics, incorporating learned relative-mobility information into generation. \pref{fig:framework} summarizes the sequence-to-ensemble task and the flexibility-calibrated bridge used by \our.

Our main contributions are:
\begin{itemize}[itemsep=5.5pt, parsep=0pt, partopsep=0pt, leftmargin=*, topsep=2pt]
    \item We formulate RNA sequence-to-ensemble generation as stochastic bridge learning on \(\mathrm{SE}(3)\), where the bridge field is trained to recover conformers from noisy frame states.
    
    \item We introduce a flexibility-calibrated heteroscedastic bridge that uses chain-normalized experimental \(\mathrm{B}\)-factor to set nucleotide-wise bridge widths during training, while providing a sequence-derived latent flexibility condition to the bridge-field network at inference.
    
    \item We show on a multi-conformer RNA benchmark that \our\ improves ensemble coverage, structural fidelity, and precision--recall balance over MSA-subsampling predictors and dedicated ensemble generators while maintaining meaningful structural diversity.
\end{itemize}


\section{Preliminaries}
\label{sec:preliminaries}

\subsection{RNA Representation and Problem Setup}
\label{sec:representation}

Given an RNA sequence $s$ of length $L$, let $\mathcal S_{\mathrm{exp}}(s)=\{G_m^\star\}_{m=1}^{M_s}$ denote the available experimental state representatives for this sequence. We do not assume access to calibrated thermodynamic populations. Our goal is to learn a conditional generator $p_\theta(G\mid s)$ whose samples $\{G^{(k)}\}_{k=1}^K$ collectively recover this observed conformational support while remaining individually plausible.

Following \texttt{\textsc{RNA-FrameFlow}}~\cite{AnandJMJHMDYHL25}, we represent the frame state of nucleotide $i$ as a rigid transformation $G_i=(x_i,R_i)\in\mathrm{SE(3)}$, where $\mathrm{SE(3)}=\mathbb{R}^3\rtimes\mathrm{SO}(3)$. Here $x_i\in\mathbb{R}^3$ is the position of the $\mathrm{C4}^{\prime}$ atom, and $R_i\in\mathrm{SO}(3)$ specifies the local orientation defined by the $\mathrm{C3}^{\prime}$--$\mathrm{C4}^{\prime}$--$\mathrm{O4}^{\prime}$ sugar geometry. We write $G=(G_1,\ldots,G_L)\in\mathrm{SE(3)}^L$ for the RNA backbone-frame state. We reconstruct all heavy-atom coordinates, including complete nucleobases, using ten torsion angles $\Phi_i\in(S^1)^{10}$, represented by sine--cosine pairs, and nucleotide-specific reference geometry (\pref{app:bridge-field-inversion}). In \our, the stochastic bridge is defined over the frame state $G$; torsions are predicted as auxiliary target-structure outputs for all-atom reconstruction.

\subsection{\texorpdfstring{$\mathrm{SE(3)}$ Flow Matching}{SE(3) Flow Matching}}
\label{sec:se3_fm}

Flow matching~\cite{lipman2022flow,liu2022flow,tong2023improving} learns a velocity field that transports a tractable prior $p_0$ to a data distribution by integrating an ordinary differential equation along a prescribed probability path. For a single nucleotide frame in $\mathrm{SE(3)}$, let $(G_0,G_1)$ be a coupled pair of prior and data states, and let $t\in[0,1]$. Conditional flow matching constructs a deterministic intermediate state by geodesic interpolation:
\begin{equation}
    x_t=(1-t)x_0+t x_1,\qquad
    R_t=R_0\exp\!\left(t\,\log(R_0^\top R_1)\right).   
\end{equation}
Differentiating this path yields closed-form conditional velocity targets
\begin{equation}
    u_t^x=\frac{x_1-x_t}{1-t},
    \qquad
    u_t^R=\frac{\log(R_t^\top R_1)^\vee}{1-t}\in\mathbb{R}^3,
    \label{eq:fm-target}
\end{equation}
where $(\cdot)^\vee$ identifies an element of $\mathfrak{so}(3)$ with its Euclidean coordinate vector. A network predicting $(\hat v_t^x,\hat v_t^R)$ is trained by tangent-space regression,
\begin{equation}
\mathcal{L}_{\mathrm{SE(3)}}(\theta)=\mathbb{E}_{t,G_0,G_1}
\left[
\frac{1}{|\mathcal R|}\sum_{i\in\mathcal R}
\left(
\lambda_x \|\hat v_{t,i}^x-u_{t,i}^x\|_2^2+\lambda_R \|\hat v_{t,i}^R-u_{t,i}^R\|_2^2
\right)
\right],
\label{eq:fm-loss}
\end{equation}
where $\mathcal R$ denotes resolved frame nucleotides and $\lambda_x,\lambda_R>0$ balance translational and rotational terms. \our\ uses this $\mathrm{SE(3)}$ velocity parameterization as a reference, but replaces the deterministic geodesic path with stochastic bridge marginals whose widths vary across nucleotides.

\subsection{RNA Flexibility and \texorpdfstring{$\mathrm{B}$}{B}-factors}
\label{sec:flexibility_background}

$\mathrm{B}$-factors provide a widely available but noisy experimental proxy for local positional uncertainty~\cite{sun2019utility}. In X-ray crystallography, they are refined as atomic displacement parameters and are related, under an ideal isotropic model, to mean-squared positional displacement~\cite{carugo2018atomic}.  In NMR-derived structures, analogous $\mathrm{B}$-factor-like quantities can be estimated from coordinate dispersion across deposited structural ensembles, although these quantities are not identical to crystallographic $\mathrm{B}$-factors~\cite{hu2021nmr,emperador2010protein}. Because $\mathrm{B}$-factors also depend on experimental modality, refinement, resolution, and static disorder, we do not treat their absolute values as calibrated thermodynamic mobilities across structures. Instead, they are best used as within-chain relative mobility signals, following prior computational studies of biomolecular flexibility~\cite{guruge2018b,kouba2024learning}.


\section{Proposed Method}
\label{sec:method}
\our\ is a sequence-conditioned stochastic-bridge generator for RNA backbone-frame ensembles on $\mathrm{SE}(3)$. Building on $\mathrm{SE}(3)$ flow matching~\cite{YimCFLSGVJBJN23}, \our\ introduces a residue-wise heteroscedastic bridge whose local width is calibrated by relative $\mathrm{B}$-factor mobility. This flexibility-calibrated bridge exposes mobile nucleotides to broader training perturbations while keeping constrained regions sharper, and the model learns to recover target conformers from these calibrated noisy frames. The bridge distribution therefore provides a source of diversity while retaining target-conformer supervision.

The input sequence $s$ is encoded once by a frozen pretrained \texttt{\textsc{AlphaFold3}} (AF3) Pairformer~\cite{AbramsonADEGPRWBB24} into single and pair representations $(\mathbf{c}^{(s)},\mathbf{c}^{(p)})$. A FlexibilityNet $\psi$ learns a sequence-derived flexibility condition grounded by $\mathrm{B}$-factor-derived mobility proxies during training, while an IPA-based bridge-field network $v_\theta$~\cite{AnandJMJHMDYHL25} predicts translational and rotational bridge velocities from the noisy state $G_t$, time $t$, AF3 representations, and this flexibility condition. The predicted fields are analytically inverted to target backbone frames and combined with torsions $\hat\Phi\in(S^1)^{10L}$ for all-atom reconstruction. At inference, \our\ uses only the input sequence and does not require experimental flexibility annotations or re-computed AF3 features under perturbed evolutionary contexts.

\subsection{\texorpdfstring{Flexibility-calibrated Heteroscedastic Bridge}{Flexibility-calibrated Heteroscedastic Bridge}}
\label{sec:flexbridge}
The heteroscedastic bridge trains the model to recover each target conformer from perturbed intermediate states. Small-noise analyses of supervised learning and denoising connect input corruption to derivative-based regularization, providing a rationale for improved local robustness~\cite{bishop1995training,alain2014regularized}. Motivated by this view, \our\ exposes flexible residues to broader perturbations while keeping constrained regions sharper. We assess the resulting robustness in the recovery-stress experiments (\pref{sec:exp-recovery-basins}); \pref{app:dynamic-stability} gives a brief discussion.

To set these widths, we use experimental \(\mathrm{B}\)-factors as within-chain mobility proxies. We therefore normalize \(\mathrm{B}\)-factors within each chain:
\begin{equation}
    \widetilde B_i = \frac{B_i}{\sum_{j=1}^L B_j/L}, \qquad i=1,\ldots,L.
    \label{eq:chain_normalized_bfactor}
\end{equation}
We rescale this proxy as \(\alpha_i=(\operatorname{clip}(\widetilde B_i,0.2,1.6)-0.2)/1.4\in[0,1]\) and set the variance multiplier \(a_i=(1+2\alpha_i)^2\). The corresponding standard-deviation multiplier is \(\sqrt{a_i}=1+2\alpha_i\), consistent with Table~\ref{tab:flexbridge_hparams}. This calibration changes the training bridge only; no \(\mathrm{B}\)-factor annotation is used at inference.

\paragraph{Translation bridge.} For translations, the calibrated bridge width can be applied directly in \(\mathbb R^3\). For a nucleotide with variance multiplier \(a\), define \(\gamma_x(t;a)=\sigma_x\sqrt{a\,t(1-t)}\), where \(\sigma_x\) is a global translational scale. The translational bridge uses the Gaussian marginal:
\begin{equation}
    q_t^x(\cdot\mid x_0,x_1) = \mathcal N\left( (1-t)x_0 + t x_1, \gamma_x^2(t;a)I_3 \right).
    \label{eq:trans_bridge}
\end{equation}
This Gaussian marginal provides a Euclidean carrier for the calibrated uncertainty. The factor \(t(1-t)\) makes the stochastic term vanish at the source and target structures, while the multiplier \(a\) determines how much intermediate uncertainty the model sees at this nucleotide during training. Thus, stochasticity is calibrated at the nucleotide level.

\paragraph{Rotation bridge.} The same calibration principle is applied to nucleotide orientations. Since rotations lie on \(\mathrm{SO}(3)\), we place the stochastic perturbation in a local logarithmic chart centered on the principal geodesic from \(R_0\) to \(R_1\). Let
\[
\Omega=\log(R_0^\top R_1)^\vee,
\qquad
\bar R_t=R_0\exp\!\left(t\Omega^\wedge\right)
\]
be the geodesic center at time \(t\). For the same variance multiplier \(a\), define \(\gamma_R(t;a)=\sigma_R\sqrt{a\,t(1-t)}\), where \(\sigma_R\) is a global rotational scale. The rotation bridge is the distribution induced by a Gaussian perturbation in the tangent coordinate around \(\bar R_t\):
\begin{equation}
    q_t^R(\cdot\mid R_0,R_1)
    = (\Phi_{\bar R_t})_{\#}\,
    \mathcal N\left(0,\gamma_R^2(t;a)I_3\right),
    \qquad
    \Phi_{\bar R_t}(\xi)=\bar R_t\exp(\xi^\wedge).
\label{eq:rot_bridge_marginal}
\end{equation}

\subsection{Learning the Flexibility Condition}
\label{sec:flexibility_conditioning}

FlexibilityNet learns an explicit sequence-derived flexibility condition. This condition is grounded by relative \(\mathrm{B}\)-factor mobility during training and guides the bridge-field network during recovery. The input sequence \(s\) is encoded once by the frozen AF3 Pairformer into \((\mathbf c^{(s)},\mathbf c^{(p)})\). FlexibilityNet \(\psi\) maps these representations to supervised flexibility logits and to the flexibility condition \(h_\psi(s)\):
\begin{equation}
(o_\psi(s),h_\psi(s))=\psi(\mathbf c^{(s)},\mathbf c^{(p)}),
\label{eq:flex_profile}
\end{equation}
where \(o_\psi(s)\in\mathbb R^{L\times K_B}\) parameterizes a \(K_B\)-bucket flexibility classifier and \(h_\psi(s)\) is passed to the bridge-field network as an additional condition.

The supervised flexibility output provides a training signal: it encourages \(h_\psi(s)\) to encode residue-wise mobility information derived from relative \(\mathrm{B}\)-factors. The classifier head anchors this condition during training, and ensemble generation uses \(h_\psi(s)\) to guide the bridge-field network.

\paragraph{Flexibility grounding.} During training, the chain-normalized \(\mathrm{B}\)-factor profile grounds the flexibility head. Let \(\mathcal V\subseteq\mathcal R\) be the supervised nucleotide set with valid \(\mathrm{B}\)-factor annotations. We discretize the chain-normalized proxy \(\widetilde B_i\) into bucket labels \(b_{\mathcal V}\) over \(\mathcal V\), and train the logits with cross entropy:
\begin{equation}
\mathcal L_\mathrm{B}
=
\operatorname{CE}\!\left(o_{\psi,\mathcal V}(s), b_{\mathcal V}\right).
\label{eq:lb}
\end{equation}
This loss aligns the learned flexibility representation with relative mobility patterns observed in training structures. It does not provide test-time \(\mathrm{B}\)-factors or manually specify flexible regions during sampling. If no valid annotations are available, we omit \(\mathcal L_\mathrm{B}\) for that example.

\subsection{Training Objective}
\label{sec:training}

The bridge construction in \pref{sec:flexbridge} gives noisy states \(G_t=(x_t,R_t)\). The IPA-based network \(v_\theta\)~\cite{AnandJMJHMDYHL25} receives \(G_t\), \(t\), \((\mathbf c^{(s)},\mathbf c^{(p)})\), and \(h_\psi(s)\), and predicts translation and rotation bridge fields \((\hat u_t^x,\hat u_t^R)\). We analytically map the predicted fields back to a target frame for supervision:
\begin{equation}
\hat x_1=x_t+(1-t)\hat u_t^x,
\qquad
\hat R_1=R_t\exp\!\left(((1-t)\hat u_t^R)^\wedge\right).
\label{eq:target_frame_recovery_main}
\end{equation}
This target-frame supervision encourages consistent recovery of the target under heteroscedastic bridge noise. It also couples the \(h_\psi(s)\)-conditioned bridge-field network to the heteroscedastic training bridge: flexible nucleotides are learned from broader intermediate perturbations, while rigid nucleotides are learned from sharper bridge neighborhoods. We define
\begin{equation}
\begin{aligned}
\mathcal L_{\mathrm{SB}}
&=
\mathbb E_{t,G_0,G_1}\!\left[
\frac{1}{(1-t)^2}
\left(
\lambda_x\,\|\hat x_1-x_1\|_{\mathbb R^3}^2
+
\lambda_R\,\left\|
\log_{R_t}(\hat R_1)
-
\log_{R_t}(R_1)
\right\|_{\mathrm{SO}(3)}^2
\right)
\right].
\end{aligned}
\label{eq:stochastic_bridge_loss}
\end{equation}
Here \(\log_R(Q)=\log(R^\top Q)^\vee\), and the norms aggregate over resolved frame nucleotides. We use \(\lambda_x=2\) and \(\lambda_R=1\), as listed in Table~\ref{tab:flexbridge_hparams}.

The full training objective combines target-frame supervision with flexibility grounding and an auxiliary structural reconstruction loss:
\begin{equation}
\mathcal L_{\our}
=
\mathcal L_{\mathrm{SB}}
+ \lambda_\mathrm{B}\,\mathcal L_\mathrm{B}
+ \lambda_\mathrm{aux}\,\mathcal L_\mathrm{aux}.
\label{eq:lbridge}
\end{equation}
The auxiliary term \(\mathcal L_\mathrm{aux}\), defined in \pref{app:bridge-field-inversion}, supervises the reconstructed geometry from \((\hat x_1,\hat R_1,\hat\Phi)\). Through this objective, \(\mathrm{B}\)-factor information affects \our\ only at training time: it calibrates the bridge distribution and grounds the flexibility representation. At inference, generation is driven by the bridge-field network conditioned on the sequence-derived embedding \(h_\psi(s)\).

\subsection{Inference}
\label{sec:inference}
To generate the ensemble given an RNA sequence, we first compute the AF3 representations. During sampling, each sample $k$ starts from an independent prior $G_0^{(k)}\sim p_0$. At each integration step, the conditional drift $\hat{u}_t^x,\hat{u}_t^R$ is predicted given the current frame state $G_t$, AF3 representations and the predicted flexibility condition. Given the predicted drift, the update is given by:
\begin{equation}
    \begin{aligned}
        x_{t+\Delta t}
        &=
        x_t + \Delta t\,\hat u_t^x + g_x(t)\sqrt{\Delta t}\,\xi_t^x,\\
        R_{t+\Delta t}
        &=
        R_t \exp\!\left( \left[ \Delta t\,\hat u_t^R + g_R(t)\sqrt{\Delta t}\,\xi_t^R \right]^\wedge \right).
    \end{aligned}
    \label{eq:sde_sampler}
\end{equation}
Here $\xi_t^x,\xi_t^R\sim\mathcal N(0,I_3)$ are independent standard Gaussian noise vectors. After the frame trajectory reaches $t=1$, the predicted torsions $\hat{\Phi}$ and nucleotide-specific reference geometry reconstruct all heavy-atom coordinates. Repeating the procedure with independent initial states and sampler noise yields the generated ensemble.


\section{Experiments}
\label{sec:experiments}

We evaluate \our\ around three questions. First, does the generated ensemble improve recovery of experimentally observed conformers? Second, do generated samples remain near observed states as ensemble diversity increases? Third, does the flexibility-calibrated bridge improve conformer-specific target-frame recovery under controlled perturbations? The first two questions evaluate final generated ensembles; the third isolates the bridge-field mechanism using the ablation variants. Metric definitions and statistical tests are given in \pref{app:ensemble-metrics} and \pref{app:statistical-analysis}.

\subsection{Dataset construction}
\label{sec:exp-dataset}

Our RNA dataset totals $1{,}301$ structures, curated from PDB~\cite{berman2000protein}, retaining chains of $20$--$300$~nt with resolution $\le 4.0$~\AA. We use a temporal cutoff of June 1, 2023 and an $80\%$ CD-HIT sequence-identity exclusion criterion between training and test sequences. The test set contains $50$ multi-conformer RNA clusters. \our\ is trained on the corresponding pre-cutoff training split.

\subsection{Baselines}
\label{sec:exp-baselines}

\begin{figure}[t]
    \centering
    \includegraphics[width=\linewidth]{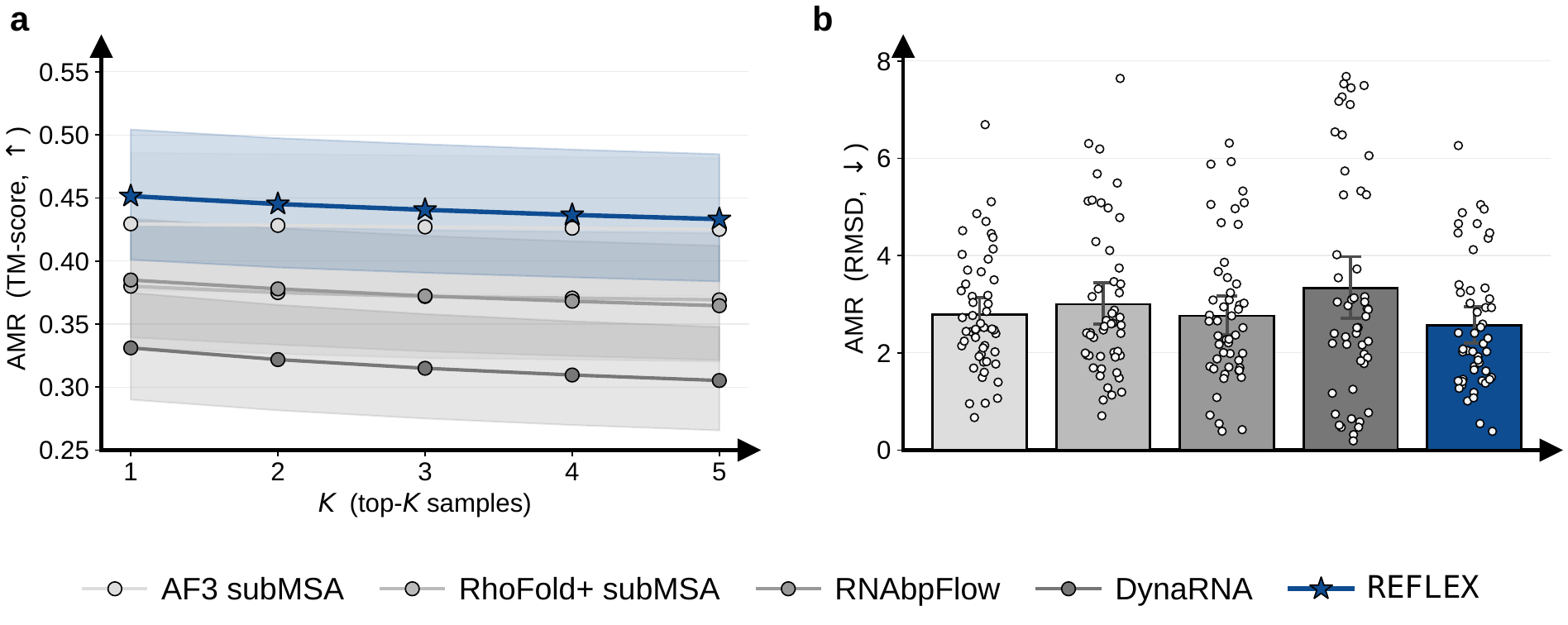}
    \caption{\textbf{Coverage of experimental conformers.} \textbf{a}: Top-$K$ AMR (TM-score), the mean of the $K$ highest matching scores for each reference conformer, for $K=1,\ldots,5$, evaluated on clusters where every method has at least five samples. \textbf{b}: Best-match RMSD AMR (\AA) on the test set. Shading and error bars show $95\%$ bootstrap confidence intervals; points in \textbf{b} show per-cluster values.}
    \label{fig:topk-amr}
    \end{figure}

We compare against two baseline families. The first uses single-structure predictors with inference-time MSA perturbations: \texttt{\textsc{AlphaFold3}}~\cite{AbramsonADEGPRWBB24} and \texttt{\textsc{RhoFold+}}~\cite{ShenHSLWWCWHXZKKWYCL24} with AFCluster-style MSA subsampling~\cite{SteeleOOAPHOCK24}. We disable templates and use the same fixed random seed for all subsampling runs. The second family consists of dedicated ensemble generators: \texttt{\textsc{RNAbpFlow}}~\cite{TarafderB25}, conditioned on sequence and \texttt{\textsc{MXFold2}}~\cite{sato2021rna} predicted base-pair information, and \texttt{\textsc{DynaRNA}}~\cite{li2025dynarna}, a coordinate-space diffusion model initialized from a randomly selected experimental conformer of the target RNA. For ensemble evaluation, all methods use a common budget of $K=20$ samples.

All external baselines use released models or checkpoints and are not retrained on our curated split. The benchmark therefore compares released systems under a common evaluation protocol; component-level attribution relies on the matched \our-family ablations.

For the recovery-stress ablations, we compare \our\ with two bridge variants trained under the same backbone architecture. Linear Bridge corresponds to the standard deterministic flow-matching path, while Uniform Stochastic Bridge uses a single global bridge width for all nucleotides and removes the flexibility head. All three models use the same recovery-stress protocol, detailed in \pref{app:ensemble-metrics}.

\subsection{\our\ improves recovery of observed conformers}
\label{sec:exp-structural-recovery}
Table~\ref{tab:precision-recall} reports full-ensemble recovery with $K=20$ samples per RNA. AMR-R averages each resolved conformer's best TM-score among generated samples; AMR-P averages each generated sample's best TM-score among resolved conformers. Both are computed per cluster and then macro-averaged. \our\ achieves the highest AMR-R ($0.451$) and AMR-P ($0.419$) among the evaluated systems, compared with $0.430$ and $0.411$ for the strongest baseline, \texttt{\textsc{AF3}} subMSA.

The paired AMR-R gain over \texttt{\textsc{AF3}} subMSA is $0.022$ (cluster-bootstrap $95\%$ CI: $[0.006,0.037]$), with higher recovery on $70\%$ of test clusters and a Holm-adjusted one-sided Wilcoxon $p=1.2\times10^{-3}$. These results support improved recovery on this benchmark.

\begin{figure}[t]
\centering
\includegraphics[width=\linewidth]{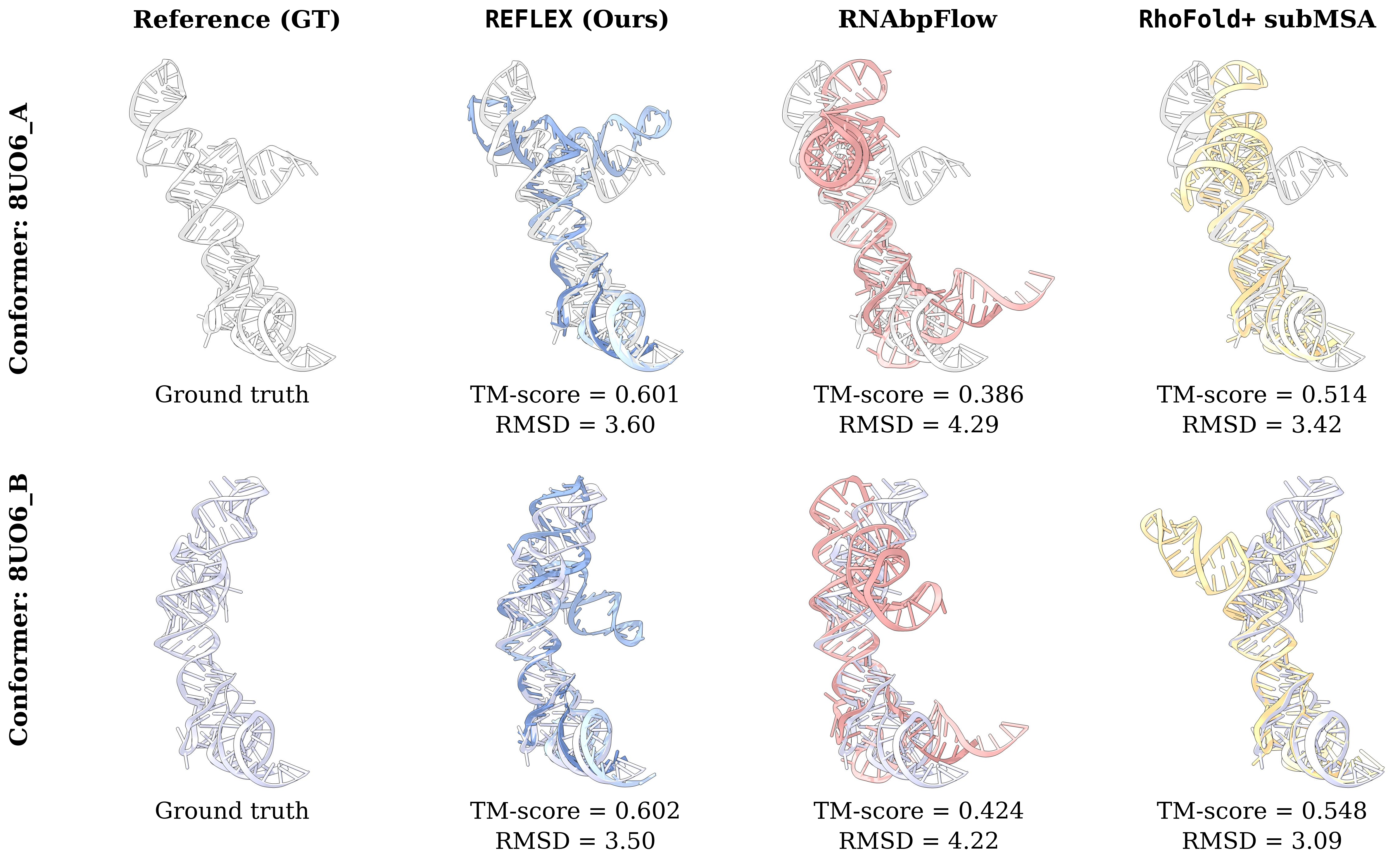}
\caption{\textbf{Qualitative recovery on PDB \texttt{8UO6}.} Two experimentally resolved conformers are shown against their nearest generated matches from \our, \texttt{\textsc{RNAbpFlow}}, and \texttt{\textsc{RhoFold+}} subMSA. TM-score and RMSD (\AA) are reported below each match.}
\label{fig:coverage-structure-comparison}
\end{figure}

Figure~\ref{fig:coverage-structure-comparison} illustrates conformer matching on PDB \texttt{8UO6}, which contains two resolved conformations. This example provides a structure-level view of generated matches; evidence for performance across targets comes from the cluster-level benchmark and paired analysis above.

At a finer structural scale, we also assess whether recovered conformers preserve experimental base-pair interactions. For each reference conformer, we annotate its highest-TM-score generated match with DSSR~\cite{lu2015dssr} and compare its base pairs with those of the reference. Matching uses only TM-score, so base-pair agreement provides a complementary measure of secondary-structure recovery. Precision, recall, and F1 are macro-averaged across test clusters; evaluation details are given in \pref{app:dssr-validation}.

\begin{table}[!htbp]
\centering
\caption{\textbf{DSSR base-pair consistency of TM-score-selected matches.} Higher values are better; the best observed value in each column is in \textbf{bold}.}
\label{tab:dssr-validation}
\begin{tabular}{@{}lccc@{}}
\toprule
Method & Precision & Recall & F1 \\
\midrule
\texttt{\textsc{AF3}} subMSA & $0.697$ & $\mathbf{0.787}$ & $0.729$ \\
\texttt{\textsc{RNAbpFlow}} & $0.712$ & $0.757$ & $0.725$ \\
\texttt{\textsc{RhoFold+}} subMSA & $0.694$ & $0.630$ & $0.647$ \\
\our & $\mathbf{0.734}$ & $0.765$ & $\mathbf{0.741}$ \\
\bottomrule
\end{tabular}
\end{table}

Table~\ref{tab:dssr-validation} shows that \our\ achieves the highest observed base-pair precision ($0.734$) and F1 ($0.741$), compared with $0.697$ and $0.729$ for \texttt{\textsc{AF3}} subMSA. Its recall ($0.765$) exceeds that of \texttt{\textsc{RNAbpFlow}} and \texttt{\textsc{RhoFold+}} subMSA, while \texttt{\textsc{AF3}} subMSA attains the highest recall ($0.787$). Together with the global recovery metrics, these results indicate that \our\ recovers conformational alternatives with competitive fidelity at the level of base-pair interactions. This evidence concerns the TM-score-selected matches and does not establish local validity across the entire generated ensemble.

\subsection{\our\ improves recovery while maintaining sample support}
\label{sec:exp-precision-recall}
Table~\ref{tab:precision-recall} evaluates whether generated ensembles place probability mass inside threshold-defined basins around experimentally observed conformers. COV-R measures how many observed conformers are recovered, while COV-P measures the complementary precision question: how many generated samples fall within one of these observed-state basins. \our\ achieves the best COV-R and COV-P among the evaluated systems at every TM-score threshold \(\tau\in\{0.30,0.35,0.40\}\). At $\tau=0.40$, its mean per-cluster recovered-reference and matched-sample fractions are $54.3\%$ and $45.0\%$, respectively. Recovery remains incomplete, so these gains do not establish exhaustive or near-native reconstruction of physical RNA ensembles.

The diversity results show that the coverage gains do not require the largest sample spread. \our\ has average pairwise RMSD of $2.900$~\AA, above \texttt{\textsc{AF3}} subMSA ($0.786$~\AA) and \texttt{\textsc{RhoFold+}} subMSA ($1.237$~\AA), but below \texttt{\textsc{RNAbpFlow}} ($3.178$~\AA) and \texttt{\textsc{DynaRNA}} ($4.439$~\AA). Its higher COV-R and COV-P across all thresholds indicate a better balance of observed-conformer recovery and sample support on this benchmark.

\begin{table}[t]
\centering
\caption{\textbf{Ensemble recovery on the RNA benchmark.} All methods generate $K=20$ samples per RNA. AMR-R and AMR-P report reference-to-sample and sample-to-reference best-match TM-scores. COV-R and COV-P use thresholds $0.30$, $0.35$, and $0.40$. Each metric is macro-averaged over its evaluated clusters (\pref{app:ensemble-metrics}); higher AMR and coverage values are better, with the best in \textbf{bold}. Diversity is average pairwise RMSD in \AA.}
\label{tab:precision-recall}
\setlength{\tabcolsep}{2pt}
\begin{tabular}{@{}lccccccccc@{}}
\toprule
\multirow{2}{*}{Method} & \multirow{2}{*}{AMR-R} & \multicolumn{3}{c}{COV-R} & \multirow{2}{*}{AMR-P} & \multicolumn{3}{c}{COV-P} & \multirow{2}{*}{Diversity (\AA)} \\
\cmidrule(lr){3-5}\cmidrule(lr){7-9}
 & & @$0.30$ & @$0.35$ & @$0.40$ & & @$0.30$ & @$0.35$ & @$0.40$ & \\
\midrule
\texttt{\textsc{RNAbpFlow}}            & $0.385$ & $0.640$ & $0.422$ & $0.358$ & $0.378$ & $0.558$ & $0.386$ & $0.371$ & $3.178$ \\
\texttt{\textsc{AF3}} subMSA           & $0.430$ & $0.646$ & $0.542$ & $0.445$ & $0.411$ & $0.563$ & $0.483$ & $0.394$ & $0.786$ \\
\texttt{\textsc{RhoFold+}} subMSA      & $0.380$ & $0.552$ & $0.470$ & $0.384$ & $0.402$ & $0.522$ & $0.437$ & $0.388$ & $1.237$ \\
\texttt{\textsc{DynaRNA}}              & $0.331$ & $0.436$ & $0.327$ & $0.247$ & $0.257$ & $0.250$ & $0.218$ & $0.204$ & $4.439$ \\
\midrule
\our                 & $\mathbf{0.451}$ & $\mathbf{0.742}$ & $\mathbf{0.647}$ & $\mathbf{0.543}$ & $\mathbf{0.419}$ & $\mathbf{0.634}$ & $\mathbf{0.533}$ & $\mathbf{0.450}$ & $2.900$ \\
\bottomrule
\end{tabular}
\end{table}

\begin{figure}[t]
\centering
\includegraphics[width=\linewidth]{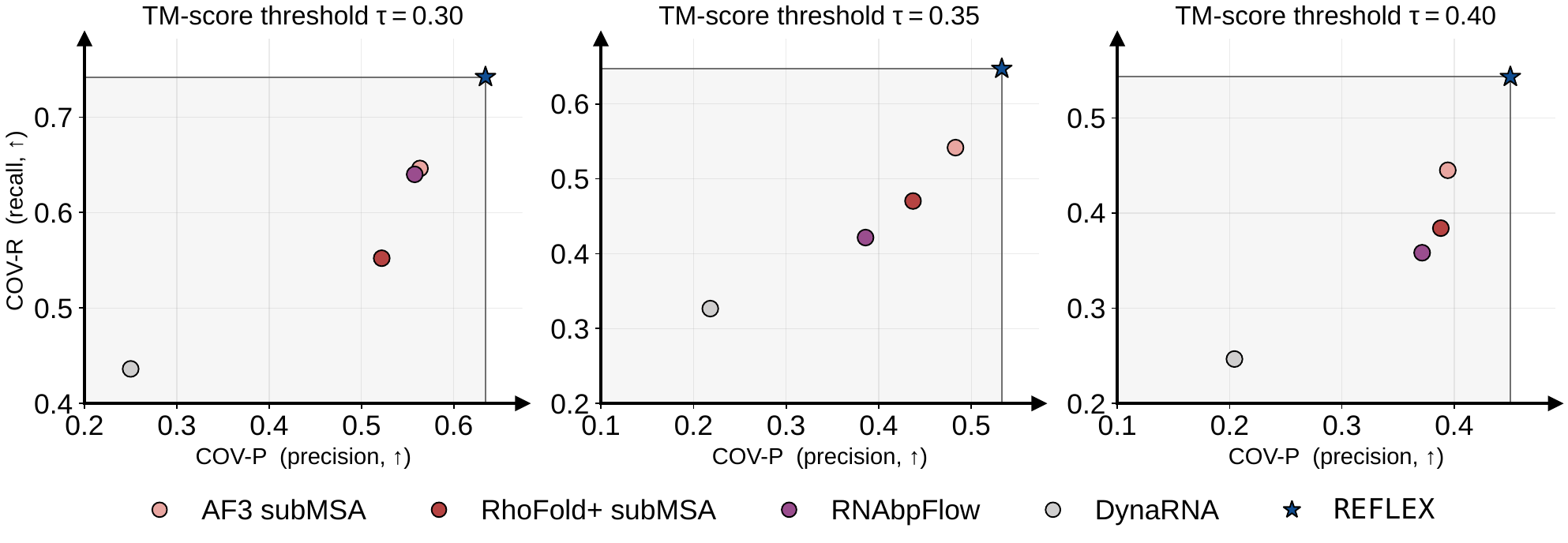}
\caption{\textbf{Precision--recall Pareto plot.} Each method is evaluated at TM-score thresholds \(\tau\in\{0.30,0.35,0.40\}\). COV-R and COV-P are macro-averaged as in Table~\ref{tab:precision-recall}; the gray shaded region denotes the dominated region.}
\label{fig:pareto}
\end{figure}

\subsection{\our\ learns more reliable conformer-specific recovery}
\label{sec:exp-recovery-basins}
\begin{figure}[!thbp]
\centering
\includegraphics[width=\linewidth]{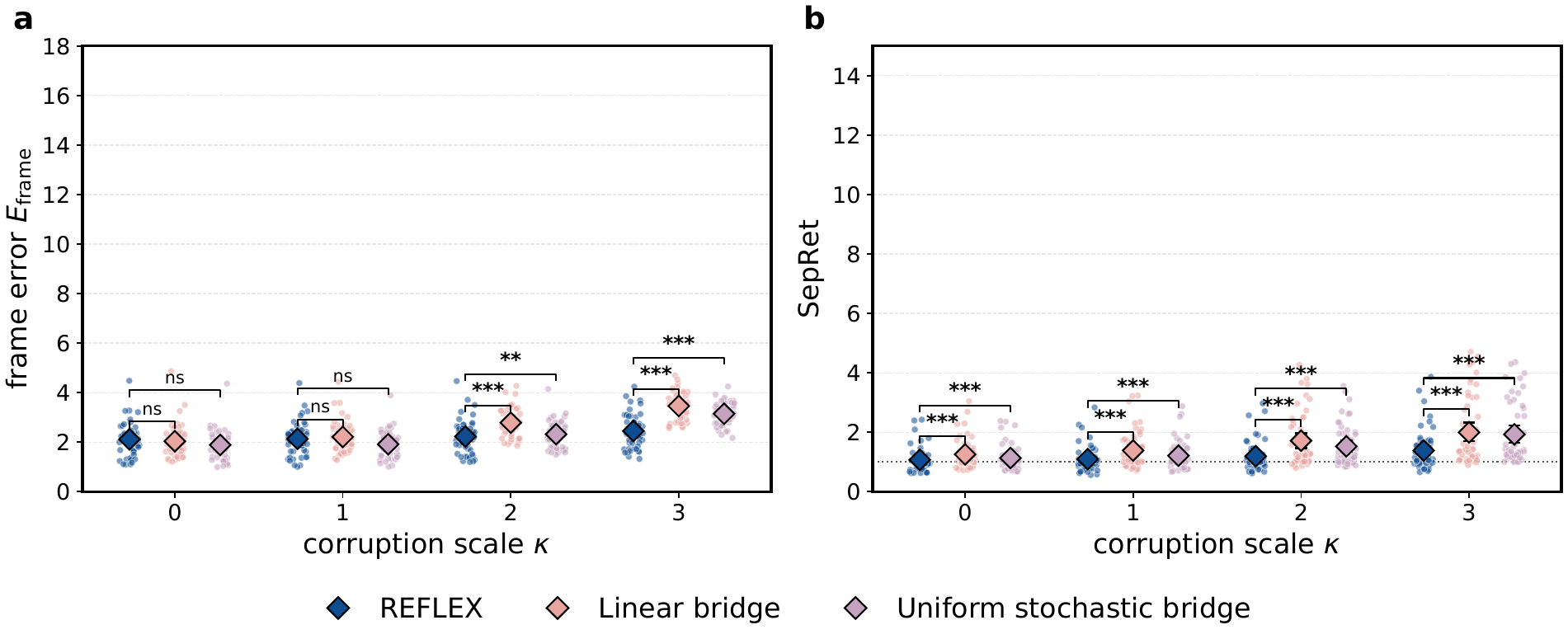}
\caption{\textbf{Multi-conformer supervision stability.} \textbf{a}: Target-frame recovery error after one-step recovery from perturbed bridge states at \(t=0.75\). \textbf{b}: SepRet under the same perturbation protocol. Stars and \texttt{ns} denote significant and non-significant paired comparisons with \our. Additional results at \(t=0.25\) and \(t=0.5\) are provided in \pref{app:additional-results}.}
\label{fig:ablation-modes}
\end{figure}

We examine conformer-specific recovery with Linear Bridge, Uniform Stochastic Bridge, and \our. For each multi-conformer sequence, we perturb bridge states around each target conformer at \(t\in\{0.25,0.5,0.75\}\) with corruption scale \(\kappa\in\{0,1,2,3\}\), then ask each model to recover the target in one step. Figure~\ref{fig:ablation-modes} shows the results at \(t=0.75\).

Figure~\ref{fig:ablation-modes} reports two complementary metrics. \(E_{\mathrm{frame}}\) measures target-frame recovery error, so lower values indicate more accurate structural recovery; SepRet measures the ratio between recovered conformer separation and ground-truth conformer separation, so values closer to \(1\) indicate better preservation of conformer separation. At \(t=0.75\), \our\ has lower mean recovery error than both alternatives at \(\kappa=2,3\), with significant paired differences. At \(\kappa=0,1\), the frame-error differences are not significant. Mean SepRet is closer to \(1\) for \our\ across all tested corruption levels, with significant paired differences from both alternatives.

Table~\ref{tab:distinctness-ablation} stratifies the test set by ground-truth conformer distinctness. Uniform Stochastic Bridge has the highest mean AMR-R in the Low bin ($0.540$ versus $0.531$ for \our), while \our\ leads in the Mid and High bins. The High-bin margin over Uniform Stochastic Bridge is only $0.002$.

The recovery-stress analysis supports improved target-frame recovery under stronger perturbations, with gains that depend on bridge time and corruption strength. Matched ablations and FlexibilityNet validation are provided in \pref{app:validation}.

\begin{table}[!thbp]
\centering
\caption{\textbf{Conformer recovery by ground-truth distinctness.} Distinctness is average pairwise TM-distance between resolved conformers. The Low/Mid/High bins contain $16/17/17$ clusters. Entries are mean per-cluster AMR-R over the evaluated structure sets; higher is better.}
\label{tab:distinctness-ablation}
\setlength{\tabcolsep}{4pt}
\begin{tabularx}{\linewidth}{@{}l *{3}{>{\centering\arraybackslash}X}@{}}
\toprule
Method & Low \([0.078, 0.363]\) & Mid \([0.365, 0.471]\) & High \([0.483, 0.843]\) \\
\midrule
Linear Bridge & $0.505$ & $0.435$ & $0.340$ \\
Uniform Stochastic Bridge & $\mathbf{0.540}$ & $0.444$ & $0.356$ \\
\our & $0.531$ & $\mathbf{0.470}$ & $\mathbf{0.358}$ \\
\bottomrule
\end{tabularx}
\end{table}

\FloatBarrier

\section{Related Work}
\label{sec:related}

\paragraph{Traditional RNA structure prediction.}
Traditional RNA tertiary-structure modeling largely relied on secondary-structure-aware motif assembly, fragment assembly, and coarse-grained or energy-based search. \texttt{\textsc{MC-Fold/MC-Sym}} uses nucleotide cyclic motifs to connect sequence-level folding with three-dimensional assembly~\cite{parisien2008mcfold}, while \texttt{\textsc{RNAComposer}} and \texttt{\textsc{3dRNA}} automate secondary-structure-guided construction from fragment or template libraries~\cite{popenda2012rnacomposer,zhao2012threedrna}. Sampling-based methods such as \texttt{\textsc{SimRNA}} and \texttt{\textsc{FARFAR2}} explore candidate folds with coarse-grained statistical potentials or Rosetta refinement~\cite{boniecki2015simrna,watkins2020farfar2}. This line of work organizes RNA 3D prediction around a pipeline of structural constraints, candidate assembly or sampling, and subsequent scoring or refinement.

\paragraph{Learning-based RNA modeling.}
Recent learning-based predictors have substantially improved RNA structure prediction. \texttt{\textsc{trRosettaRNA}}, \texttt{\textsc{RhoFold+}}, \texttt{\textsc{NuFold}}, and \texttt{\textsc{AlphaFold3}} predict accurate representative structures using learned sequence, evolutionary, or geometric representations~\cite{wang2023trrosettarna,ShenHSLWWCWHXZKKWYCL24,kagaya2025nufold,AbramsonADEGPRWBB24}. This family of methods shifts RNA structure prediction from hand-designed assembly and search toward learned structural priors trained on available RNA structures. Alternative conformations can be obtained by inference with heuristically subsampled MSAs, which probes conformational variation by changing the evolutionary context supplied to a structure predictor~\cite{SteeleOOAPHOCK24}.

RNA inverse design and co-design have also been explored through reinforcement-learning-guided diffusion, joint topology-and-geometry objectives, and cooperative multi-agent learning~\cite{hu2026rider,hu2026tango,hu2026cocorna}. These approaches address RNA design, complementing the sequence-conditioned conformational ensemble modeling studied here.

\paragraph{Flow matching models.}
Flow matching learns a continuous path that transports samples from a tractable prior to the data distribution~\cite{lipman2022flow,liu2022flow,tong2023improving,albergo2025stochastic,song2020score,shi2023diffusion,tong2023simulation}. For molecular structures, geometric generative models define such dynamics on Euclidean coordinates, rotations, or \(\mathrm{SE(3)}\) frames~\cite{yim2023se,bose2023se,YimCFLSGVJBJN23}. RNA-specific generative models have adapted these ideas to backbone-frame generation, base-pair-augmented backbone generation, and diffusion-based conformational modeling~\cite{AnandJMJHMDYHL25,TarafderB25,li2025dynarna}.

\section{Conclusion}
\label{sec:conclusion}
We presented \our, a sequence-conditioned RNA ensemble generator that models conformational diversity as a nucleotide-wise property of the molecule. The method learns a flexibility-calibrated heteroscedastic stochastic bridge on \(\mathrm{SE}(3)\), using chain-normalized experimental \(\mathrm{B}\)-factors during training to set local stochastic strength and a sequence-derived flexibility condition at inference. On the test set, \our\ better recovers experimentally observed conformational states, places more generated structures inside observed-state basins, and achieves the strongest precision--recall trade-off while maintaining meaningful diversity. The recovery-stress ablation indicates that target-frame recovery gains depend on bridge time and corruption strength: at \(t=0.75\), recovery error is significantly lower under stronger perturbations, while differences at lower corruption levels are not significant. Mean separation retention is closer to the reference value across the tested corruption levels. These results support the central claim that RNA ensemble generation benefits from calibrating where stochasticity is placed, providing a route toward sequence-conditioned models that capture both global alternative states and local flexibility patterns.

\section*{Acknowledgements}
This work was supported by the UKRI Future Leaders Fellowship under grants MR/S017062/1 and MR/X011135/1; the National Natural Science Foundation of China under grants 62376056 and 62076056; Isambard AI Open Access (AIOA - 01 - 2026-06-04 - EvoRNA: Predictive Scaling of Evolution-Aware and Multimodal RNA Foundation Models for Crop Resilience); GPU compute credits provided through the Modal for Academics program; the Royal Society Faraday Discovery Fellowship (FDF/S2/251014); BBSRC Transformative Research Technologies (UKRI1875); the Royal Society International Exchanges Award (IES/R3/243136); the Kan Tong Po Fellowship (KTP/R1/231017); the Amazon Research Award; and the Alan Turing Fellowship.

\bibliographystyle{unsrtnat}
\bibliography{flexbridge}

@article{sato2021rna,
  title={{RNA} secondary structure prediction using deep learning with thermodynamic integration},
  author={Sato, Kengo and Akiyama, Manato and Sakakibara, Yasubumi},
  journal={Nature communications},
  volume={12},
  number={1},
  pages={941},
  year={2021},
  publisher={Nature Publishing Group UK London}
}

@article{AbramsonADEGPRWBB24,
  author    = {Josh Abramson and
               Jonas Adler and
               Jack Dunger and
               Richard Evans and
               Tim Green and
               Alexander Pritzel and
               Olaf Ronneberger and
               Lindsay Willmore and
               Andrew J. Ballard and
               Joshua Bambrick and
               others},
  title     = {Accurate structure prediction of biomolecular interactions with {AlphaFold3}},
  journal   = {Nature},
  volume    = {630},
  number    = {8016},
  pages     = {493--500},
  year      = {2024},
  publisher = {Nature Publishing Group UK London}
}

@article{ShenHSLWWCWHXZKKWYCL24,
  author    = {Tao Shen and
               Zhihang Hu and
               Siqi Sun and
               Di Liu and
               Felix Wong and
               Jiuming Wang and
               Jiayang Chen and
               Yixuan Wang and
               Liang Hong and
               Jin Xiao and
               Liangzhen Zheng and
               Tejas Krishnamoorthi and
               Irwin King and
               Sheng Wang and
               Peng Yin and
               James J. Collins and
               Yu Li},
  title     = {Accurate {RNA 3D} structure prediction using a language model-based deep learning approach},
  journal   = {Nat. Methods},
  volume    = {21},
  number    = {12},
  pages     = {2287--2298},
  year      = {2024},
  publisher = {Nature Publishing Group US New York}
}

@article{TarafderB25,
  author  = {Sumit Tarafder and
             Debswapna Bhattacharya},
  title   = {{RNAbpFlow}: Base pair-augmented {SE}(3)-flow matching for conditional {RNA 3D} structure generation},
  journal = {bioRxiv},
  year    = {2025}
}

@article{SteeleOOAPHOCK24,
  author    = {Hannah K. Wayment-Steele and
               Adedolapo Ojoawo and
               Renee Otten and
               Julia M. Apitz and
               Warintra Pitsawong and
               Marc H{\"o}mberger and
               Sergey Ovchinnikov and
               Lucy Colwell and
               Dorothee Kern},
  title     = {Predicting multiple conformations via sequence clustering and {AlphaFold2}},
  journal   = {Nature},
  volume    = {625},
  number    = {7996},
  pages     = {832--839},
  year      = {2024},
  publisher = {Nature Publishing Group UK London}
}

@article{AnandJMJHMDYHL25,
  author       = {Rishabh Anand and
                  Chaitanya K. Joshi and
                  Alex Morehead and
                  Arian Rokkum Jamasb and
                  Charles Harris and
                  Simon V. Mathis and
                  Kieran Didi and
                  Rex Ying and
                  Bryan Hooi and
                  Pietro Lio},
  title        = {{RNA-FrameFlow}: Flow Matching for de novo {3D RNA} Backbone Design},
  journal      = {Trans. Mach. Learn. Res.},
  volume       = {2025},
  year         = {2025},
  bibsource    = {dblp computer science bibliography, https://dblp.org}
}

@inproceedings{YimCFLSGVJBJN23,
  author    = {Jason Yim and
               Andrew Campbell and
               Andrew Y. K. Foong and
               Michael Gastegger and
               Jos{\'e} Jim{\'e}nez-Luna and
               Sarah Lewis and
               Victor Garcia Satorras and
               Bastiaan S. Veeling and
               Regina Barzilay and
               Tommi Jaakkola and
               Frank No{\'e}},
  title     = {Fast protein backbone generation with {SE(3)} flow matching},
  booktitle = {NeurIPS 2023 Machine Learning in Structural Biology Workshop},
  year      = {2023}
}

@article{lipman2022flow,
  title   = {Flow matching for generative modeling},
  author  = {Lipman, Yaron and
             Chen, Ricky TQ and
             Ben-Hamu, Heli and
             Nickel, Maximilian and
             Le, Matt},
  journal = {arXiv preprint arXiv:2210.02747},
  year    = {2022}
}

@article{liu2022flow,
  title   = {Flow straight and fast: Learning to generate and transfer data with rectified flow},
  author  = {Liu, Xingchao and
             Gong, Chengyue and
             Liu, Qiang},
  journal = {arXiv preprint arXiv:2209.03003},
  year    = {2022}
}

@article{bose2023se,
  title   = {{SE}(3)-stochastic flow matching for protein backbone generation},
  author  = {Bose, Avishek Joey and
             Akhound-Sadegh, Tara and
             Huguet, Guillaume and
             Fatras, Kilian and
             Rector-Brooks, Jarrid and
             Liu, Cheng-Hao and
             Nica, Andrei Cristian and
             Korablyov, Maksym and
             Bronstein, Michael and
             Tong, Alexander},
  journal = {arXiv preprint arXiv:2310.02391},
  year    = {2023}
}

@inproceedings{yim2023se,
  title     = {{SE}(3) diffusion model with application to protein backbone generation},
  author    = {Yim, Jason and
               Trippe, Brian L and
               De Bortoli, Valentin and
               Mathieu, Emile and
               Doucet, Arnaud and
               Barzilay, Regina and
               Jaakkola, Tommi},
  booktitle = {International Conference on Machine Learning (ICML)},
  year      = {2023}
}

@article{tong2023improving,
  title   = {Improving and generalizing flow-based generative models with minibatch optimal transport},
  author  = {Tong, Alexander and
             Fatras, Kilian and
             Malkin, Nikolay and
             Huguet, Guillaume and
             Zhang, Yanlei and
             Rector-Brooks, Jarrid and
             Wolf, Guy and
             Bengio, Yoshua},
  journal = {arXiv preprint arXiv:2302.00482},
  year    = {2023}
}

@article{albergo2025stochastic,
  title   = {Stochastic interpolants: A unifying framework for flows and diffusions},
  author  = {Albergo, Michael and
             Boffi, Nicholas M and
             Vanden-Eijnden, Eric},
  journal = {Journal of Machine Learning Research},
  volume  = {26},
  number  = {209},
  pages   = {1--80},
  year    = {2025}
}

@article{shi2023diffusion,
  title   = {Diffusion {Schr{\"o}dinger} bridge matching},
  author  = {Shi, Yuyang and
             De Bortoli, Valentin and
             Campbell, Andrew and
             Doucet, Arnaud},
  journal = {Advances in neural information processing systems},
  volume  = {36},
  pages   = {62183--62223},
  year    = {2023}
}

@article{tong2023simulation,
  title   = {Simulation-free {Schr{\"o}dinger} bridges via score and flow matching},
  author  = {Tong, Alexander and
             Malkin, Nikolay and
             Fatras, Kilian and
             Atanackovic, Lazar and
             Zhang, Yanlei and
             Huguet, Guillaume and
             Wolf, Guy and
             Bengio, Yoshua},
  journal = {arXiv preprint arXiv:2307.03672},
  year    = {2023}
}

@article{song2020score,
  title   = {Score-based generative modeling through stochastic differential equations},
  author  = {Song, Yang and
             Sohl-Dickstein, Jascha and
             Kingma, Diederik P and
             Kumar, Abhishek and
             Ermon, Stefano and
             Poole, Ben},
  journal = {arXiv preprint arXiv:2011.13456},
  year    = {2020}
}

@article{mustoe2014hierarchy,
  title     = {Hierarchy of {RNA} functional dynamics},
  author    = {Mustoe, Anthony M and
               Brooks, Charles L and
               Al-Hashimi, Hashim M},
  journal   = {Annual review of biochemistry},
  volume    = {83},
  number    = {1},
  pages     = {441--466},
  year      = {2014},
  publisher = {Annual Reviews}
}

@article{maruvsivc2019rna,
  title     = {{RNA} dynamics by {NMR} spectroscopy},
  author    = {Maru{\v{s}}i{\v{c}}, Maja and
               Schlagnitweit, Judith and
               Petzold, Katja},
  journal   = {ChemBioChem},
  volume    = {20},
  number    = {21},
  pages     = {2685--2710},
  year      = {2019},
  publisher = {Wiley Online Library}
}

@article{carugo2018atomic,
  title     = {Atomic displacement parameters in structural biology},
  author    = {Carugo, Oliviero},
  journal   = {Amino Acids},
  volume    = {50},
  number    = {7},
  pages     = {775--786},
  year      = {2018},
  publisher = {Springer}
}

@article{guruge2018b,
  title     = {{B-factor} profile prediction for {RNA} flexibility using support vector machines},
  author    = {Guruge, Ivantha and
               Taherzadeh, Ghazaleh and
               Zhan, Jian and
               Zhou, Yaoqi and
               Yang, Yuedong},
  journal   = {Journal of Computational Chemistry},
  volume    = {39},
  number    = {8},
  pages     = {407--411},
  year      = {2018},
  publisher = {Wiley Online Library}
}

@article{ganser2019roles,
  title     = {The roles of structural dynamics in the cellular functions of {RNAs}},
  author    = {Ganser, Laura R and
               Kelly, Megan L and
               Herschlag, Daniel and
               Al-Hashimi, Hashim M},
  journal   = {Nature reviews Molecular cell biology},
  volume    = {20},
  number    = {8},
  pages     = {474--489},
  year      = {2019},
  publisher = {Nature Publishing Group UK London}
}

@article{li2025dynarna,
  title     = {{DynaRNA}: accurate dynamic {RNA} conformation ensemble generation with diffusion model},
  author    = {Li, Zhengxin and
               Zhu, Junjie and
               Hong, Xiaokun and
               Mu, Junxi and
               Zheng, Zhuoqi and
               Cui, Taeyoung and
               Sun, Yutong and
               Wei, Ting and
               Chen, Hai-Feng},
  journal   = {Communications Biology},
  volume    = {8},
  number    = {1},
  pages     = {1472},
  year      = {2025},
  publisher = {Nature Publishing Group UK London}
}

@article{berman2000protein,
  title     = {The protein data bank},
  author    = {Berman, Helen M and
               Westbrook, John and
               Feng, Zukang and
               Gilliland, Gary and
               Bhat, Talapady N and
               Weissig, Helge and
               Shindyalov, Ilya N and
               Bourne, Philip E},
  journal   = {Nucleic acids research},
  volume    = {28},
  number    = {1},
  pages     = {235--242},
  year      = {2000},
  publisher = {Oxford University Press}
}

@article{ken2023rna,
  title     = {{RNA} conformational propensities determine cellular activity},
  author    = {Ken, Megan L and
               Roy, Rohit and
               Geng, Ainan and
               Ganser, Laura R and
               Manghrani, Akanksha and
               Cullen, Bryan R and
               Schulze-Gahmen, Ursula and
               Herschlag, Daniel and
               Al-Hashimi, Hashim M},
  journal   = {Nature},
  volume    = {617},
  number    = {7962},
  pages     = {835--841},
  year      = {2023},
  publisher = {Nature Publishing Group UK London}
}

@article{kagaya2025nufold,
  title     = {{NuFold}: end-to-end approach for {RNA} tertiary structure prediction with flexible nucleobase center representation},
  author    = {Kagaya, Yuki and
               Zhang, Zicong and
               Ibtehaz, Nabil and
               Wang, Xiao and
               Nakamura, Tsukasa and
               Punuru, Pranav Deep and
               Kihara, Daisuke},
  journal   = {Nature communications},
  volume    = {16},
  number    = {1},
  pages     = {881},
  year      = {2025},
  publisher = {Nature Publishing Group UK London}
}

@article{sun2019utility,
  title     = {Utility of {B-factors} in protein science: interpreting rigidity, flexibility, and internal motion and engineering thermostability},
  author    = {Sun, Zhoutong and
               Liu, Qian and
               Qu, Ge and
               Feng, Yan and
               Reetz, Manfred T},
  journal   = {Chemical reviews},
  volume    = {119},
  number    = {3},
  pages     = {1626--1665},
  year      = {2019},
  publisher = {ACS Publications}
}

@article{kouba2024learning,
  title   = {Learning to engineer protein flexibility},
  author  = {Kouba, Petr and
             Planas-Iglesias, Joan and
             Damborsky, Jiri and
             Sedlar, Jiri and
             Mazurenko, Stanislav and
             Sivic, Josef},
  journal = {arXiv preprint arXiv:2412.18275},
  year    = {2024}
}

@article{hu2021nmr,
  title     = {{NMR}-based methods for protein analysis},
  author    = {Hu, Yunfei and
               Cheng, Kai and
               He, Lichun and
               Zhang, Xu and
               Jiang, Bin and
               Jiang, Ling and
               Li, Conggang and
               Wang, Guan and
               Yang, Yunhuang and
               Liu, Maili},
  journal   = {Analytical chemistry},
  volume    = {93},
  number    = {4},
  pages     = {1866--1879},
  year      = {2021},
  publisher = {ACS Publications}
}

@article{emperador2010protein,
  title     = {Protein flexibility from discrete molecular dynamics simulations using quasi-physical potentials},
  author    = {Emperador, Agust{\'\i} and
               Meyer, Tim and
               Orozco, Modesto},
  journal   = {Proteins: Structure, Function, and Bioinformatics},
  volume    = {78},
  number    = {1},
  pages     = {83--94},
  year      = {2010},
  publisher = {Wiley Online Library}
}

@article{lu2015dssr,
  title     = {{DSSR}: an integrated software tool for dissecting the spatial structure of {RNA}},
  author    = {Lu, Xiang-Jun and Bussemaker, Harmen J. and Olson, Wilma K.},
  journal   = {Nucleic Acids Research},
  volume    = {43},
  number    = {21},
  pages     = {e142},
  year      = {2015},
  doi       = {10.1093/nar/gkv716}
}

@article{parisien2008mcfold,
  title     = {The {MC-Fold} and {MC-Sym} pipeline infers {RNA} structure from sequence data},
  author    = {Parisien, Marc and Major, Fran{\c{c}}ois},
  journal   = {Nature},
  volume    = {452},
  number    = {7183},
  pages     = {51--55},
  year      = {2008},
  publisher = {Springer Science and Business Media LLC},
  doi       = {10.1038/nature06684}
}

@article{popenda2012rnacomposer,
  title     = {Automated 3{D} structure composition for large {RNAs}},
  author    = {Popenda, Mariusz and Szachniuk, Marta and Antczak, Maciej and Purzycka, Katarzyna J. and Lukasiak, Piotr and Bartol, Natalia and Blazewicz, Jacek and Adamiak, Ryszard W.},
  journal   = {Nucleic Acids Research},
  volume    = {40},
  number    = {14},
  pages     = {e112--e112},
  year      = {2012},
  publisher = {Oxford University Press},
  doi       = {10.1093/nar/gks339}
}

@article{zhao2012threedrna,
  title     = {Automated and fast building of three-dimensional {RNA} structures},
  author    = {Zhao, Yunjie and Huang, Yangyu and Gong, Zhou and Wang, Yanjie and Man, Jianfen and Xiao, Yi},
  journal   = {Scientific Reports},
  volume    = {2},
  number    = {1},
  year      = {2012},
  publisher = {Springer Science and Business Media LLC},
  doi       = {10.1038/srep00734}
}

@article{boniecki2015simrna,
  title     = {{SimRNA}: a coarse-grained method for {RNA} folding simulations and 3{D} structure prediction},
  author    = {Boniecki, Michal J. and Lach, Grzegorz and Dawson, Wayne K. and Tomala, Konrad and Lukasz, Pawel and Soltysinski, Tomasz and Rother, Kristian M. and Bujnicki, Janusz M.},
  journal   = {Nucleic Acids Research},
  volume    = {44},
  number    = {7},
  pages     = {e63},
  year      = {2016},
  publisher = {Oxford University Press},
  doi       = {10.1093/nar/gkv1479}
}

@article{watkins2020farfar2,
  title     = {{FARFAR2}: Improved de novo {Rosetta} prediction of complex global {RNA} folds},
  author    = {Watkins, Andrew Martin and Rangan, Ramya and Das, Rhiju},
  journal   = {Structure},
  volume    = {28},
  number    = {8},
  pages     = {963--976.e6},
  year      = {2020},
  publisher = {Elsevier},
  doi       = {10.1016/j.str.2020.05.011}
}

@article{bishop1995training,
  title     = {Training with Noise is Equivalent to {Tikhonov} Regularization},
  author    = {Bishop, Christopher M.},
  journal   = {Neural Computation},
  volume    = {7},
  number    = {1},
  pages     = {108--116},
  year      = {1995},
  doi       = {10.1162/neco.1995.7.1.108}
}

@article{alain2014regularized,
  title     = {What Regularized Auto-Encoders Learn from the Data-Generating Distribution},
  author    = {Alain, Guillaume and Bengio, Yoshua},
  journal   = {Journal of Machine Learning Research},
  volume    = {15},
  number    = {110},
  pages     = {3743--3773},
  year      = {2014},
}

@article{wang2023trrosettarna,
  title     = {{trRosettaRNA}: automated prediction of {RNA} 3{D} structure with transformer network},
  author    = {Wang, Wenkai and Feng, Chenjie and Han, Renmin and Wang, Ziyi and Ye, Lisha and Du, Zongyang and Wei, Hong and Zhang, Fa and Peng, Zhenling and Yang, Jianyi},
  journal   = {Nature Communications},
  volume    = {14},
  number    = {1},
  year      = {2023},
  publisher = {Springer Science and Business Media LLC},
  doi       = {10.1038/s41467-023-42528-4}
}

@inproceedings{hu2026rider,
  author    = {Hu, Tianmeng and Cui, Yongzheng and Luo, Biao and Li, Ke},
  title     = {{RIDER}: {3D} {RNA} Inverse Design with Reinforcement Learning-Guided Diffusion},
  booktitle = {Proc. International Conference on Learning Representations},
  year      = {2026},
  volume    = {2026},
  pages     = {79321--79346},
}

@inproceedings{hu2026cocorna,
  author    = {Hu, Tianmeng and Luo, Biao and Li, Ke},
  title     = {{CocoRNA}: Collective {RNA} Design with Cooperative Multi-agent Reinforcement Learning},
  booktitle = {Proc. International Conference on Machine Learning},
  year      = {2026},
  volume    = {306},
  pages     = {45495--45519},
}

@misc{wang2026separatingcommonuniquedirections,
  title         = {Separating Common and Unique Directions for Model Merging},
  author        = {Wang, Shiting and Zou, Yingjie and Li, Ke},
  year          = {2026},
  eprint        = {2611.04108},
  archivePrefix = {arXiv},
  primaryClass  = {cs.LG},
}

@inproceedings{li2026understandingtransfer,
  title     = {Understanding Transfer Learning of {RNA} Foundation Models on Downstream Tasks},
  author    = {Li, Yuan and Yang, Heng and Chen, Renzhi and Li, Ke},
  booktitle = {Proceedings of the 43rd International Conference on Machine Learning},
  volume    = {306},
  pages     = {71724--71744},
  year      = {2026},
  publisher = {PMLR},
}

@article{zhou2026factchecking,
  title   = {Assessing Automated Fact-Checking for Medical {LLM} Responses with Knowledge Graphs},
  author  = {Zhou, Shasha and Huang, Mingyu and Cole, Jack and Britton, Charles and Yin, Ming and Wolber, Jan and Li, Ke},
  journal = {Proceedings of the AAAI Conference on Artificial Intelligence},
  volume  = {40},
  number  = {46},
  pages   = {39646--39656},
  year    = {2026},
  doi     = {10.1609/aaai.v40i46.41317},
}

@inproceedings{zhou2026genomicinterpretability,
  title     = {Position: Genomic Model Research Must Move Beyond Anecdotal Evaluation of Interpretability Methods},
  author    = {Zhou, Shasha and Huang, Mingyu and Li, Ke},
  booktitle = {Proceedings of the 43rd International Conference on Machine Learning},
  volume    = {306},
  pages     = {172974--172999},
  year      = {2026},
  publisher = {PMLR},
}

@article{hu2026tango,
  author  = {Hu, Tianmeng and Luo, Biao and Li, Ke},
  title   = {{TANGO}: {RNA} Topology and Geometry Co-Design},
  journal = {arXiv preprint arXiv:2610.20277},
  year    = {2026},
}

@article{cole2025phyloaug,
  title     = {{PhyloAug}: An Evolutionary and Structure-Aware Data Augmentation Tool},
  author    = {Cole, Jack and Yang, Heng and Tsaneva-Atanasova, Krasimira and Li, Ke},
  journal   = {bioRxiv},
  pages     = {2025--09},
  year      = {2025},
  publisher = {Cold Spring Harbor Laboratory}
}

\clearpage
\appendix
\section{Hyperparameter Settings}
\label{app:implementation}

\our\ was trained on a single NVIDIA GH200 GPU. Table~\ref{tab:flexbridge_hparams} summarizes the main hyperparameters.

\begin{table}[H]
\centering
\caption{Key hyperparameters of \our.}
\label{tab:flexbridge_hparams}
\setlength{\tabcolsep}{4pt}
\renewcommand{\arraystretch}{1.15}
\begin{tabularx}{\linewidth}{@{}>{\raggedright\arraybackslash}p{2.7cm} >{\raggedright\arraybackslash}p{5.5cm} >{\raggedright\arraybackslash}X@{}}
\toprule
\textbf{Group} & \textbf{Symbol / Name} & \textbf{Value} \\
\midrule
\multirow{2}{=}{Time grid}
 & Cutoff $\epsilon_t$ & $10^{-2}$ \\
 & Inference / validation steps & $60$ / $60$ \\
\midrule
\multirow{4}{=}{Stochastic bridge}
 & Base widths $\sigma_x$ / $\sigma_R$ & $1.0$~\AA{} / $1.0$~rad \\
 & Heteroscedastic gain & $\sigma_i=\sigma(1+2\alpha_i)$, with $\alpha_i=(\operatorname{clip}(\widetilde B_i,0.2,1.6)-0.2)/1.4$ \\
 & Diffusion $g_x$ / $g_R$ & $1.0$~\AA{} / $1.0$~rad; constant \\
 & Rotation schedule & Exponential, $\rho=10$ \\
\midrule
\multirow{5}{=}{FlexibilityNet}
 & AF3 single input / hidden dimension & $384$ / $64$ \\
 & Transformer layers / heads / FFN & $2$ / $4$ / $256$ \\
 & Flexibility bins $K_B$ & $64$ over $\widetilde B\in[0.2,1.6]$ \\
 & Condition $h_\psi(s)$ & Bin probabilities; FiLM on single embedding \\
 & Training status & Jointly trained with $\mathcal L_{\mathrm B}$ \\
\midrule
\multirow{5}{=}{Flow model}
 & Node / edge embedding $c_s$ / $c_z$ & $256$ / $128$ \\
 & IPA blocks / heads / hidden dimension & $4$ / $8$ / $128$ \\
 & IPA query--key / value points & $8$ / $12$ \\
 & Sequence transformer heads / layers & $4$ / $2$ \\
 & Predicted torsions & $10$ ($8$ backbone + $\mathrm{O2}^{\prime}$ + $\chi$) \\
\midrule
\multirow{3}{=}{Flow-branch loss}
 & Translation / rotation weight & $2.0$ / $1.0$ \\
 & Auxiliary weight / time threshold & $1.0$ / $0.25$ \\
 & Torsion loss weight & $1.0$ \\
\midrule
\multirow{6}{=}{Optimization}
 & Optimizer & AdamW; constant learning rate \\
 & Learning rate / weight decay & $10^{-4}$ / $10^{-2}$ \\
 & Batch size & $24$ \\
 & Self-conditioning probability & $0.5$ \\
 & Gradient clipping & $1.0$, global $L_2$ norm \\
 & Training length & $400$ epochs / $22{,}000$ steps \\
\bottomrule
\end{tabularx}
\end{table}

\section{Noise Calibration and Robust Recovery}
\label{app:dynamic-stability}

Small-noise training can regularize local sensitivity~\cite{bishop1995training,alain2014regularized}. Under the regularity assumptions of Alain and Bengio~\cite{alain2014regularized}, a Euclidean denoiser \(r_\eta\) with \(r_\eta(x)=x+o(1)\) as \(\eta\to0\) satisfies
\[
\begin{aligned}
\mathcal L_{\mathrm{denoise}}(\eta)
&=\mathbb E_{x,z}\bigl\|r_\eta(x+\eta z)-x\bigr\|_2^2\\
&=\mathbb E_x\bigl\|r_\eta(x)-x\bigr\|_2^2
+\eta^2\mathbb E_x\bigl\|J_{r_\eta}(x)\bigr\|_F^2+o(\eta^2).
\end{aligned}
\]
Here \(x\) is a clean input, \(z\sim\mathcal N(0,I)\) is independent noise, and \(J_{r_\eta}\) is the denoiser's Jacobian. The \(\eta^2\)-weighted term penalizes local sensitivity to input perturbations.

This denoising result motivates our bridge design. In \our, varying bridge noise keeps the target conformer fixed, encouraging recovery over a local neighborhood. Flexibility calibration widens this neighborhood at mobile residues while keeping constrained regions sharper. We use bounded noise scales to retain target-specific information, and assess robustness empirically in \pref{sec:exp-recovery-basins}.

\section{Ensemble Evaluation Metrics}
\label{app:ensemble-metrics}

For a cluster with \(K\) generated samples and \(K_c\) experimentally resolved states, let \(\mathcal G=\{g_j\}_{j=1}^{K}\) denote the generated ensemble and \(\mathcal Y=\{y_k\}_{k=1}^{K_c}\) the experimental ensemble. These metrics test three desiderata for ensemble generation: recovering experimentally observed conformational states, avoiding unsupported generated structures, and maintaining non-collapsed ensemble spread.

\paragraph{Fidelity.}
We measure recovery of experimentally observed states using average matched recovery. For each reference conformer \(y\), order the TM-scores of all \(K\) generated samples as \(s_{(1)}(y)\ge\cdots\ge s_{(K)}(y)\). Figure~\ref{fig:topk-amr}a reports top-\(k\) AMR for \(k=1,\ldots,5\):
\[
\mathrm{AMR}_{\mathrm{TM}}\text{@}k
\;=\;
\frac{1}{K_c}\sum_{y\in\mathcal Y}
\frac{1}{k}\sum_{r=1}^{k}s_{(r)}(y)
\quad \uparrow .
\]
Here \(k\) counts the highest-scoring matches averaged within the fixed generated ensemble. The best-match case gives \(\mathrm{AMR\text{-}R}=\mathrm{AMR}_{\mathrm{TM}}\text{@}1\), as reported in Table~\ref{tab:precision-recall}. Figure~\ref{fig:topk-amr}b reports best-match RMSD AMR over the same generated ensemble:
\[
\mathrm{AMR}_{\mathrm{RMSD}}
\;=\;
\frac{1}{K_c}\sum_{y\in\mathcal Y}
\min_{g\in\mathcal G}\,\mathrm{RMSD}(g,\,y)
\quad \downarrow .
\]
The complementary sample-to-reference score is
\[
\mathrm{AMR\text{-}P}
\;=\;
\frac{1}{K}\sum_{g\in\mathcal G}
\max_{y\in\mathcal Y}\,\mathrm{TM}(g,\,y)
\quad \uparrow .
\]
We evaluate AMR-R, AMR-P, COV-R, and COV-P on the RNA benchmark. Each metric is computed per cluster and macro-averaged over its evaluated clusters.

\paragraph{Precision--recall trade-off.}
At TM-score threshold \(\tau\), recall coverage measures how many experimental states are covered by the generated ensemble, while precision coverage measures how many generated samples remain near the experimental ensemble:
\[
\mathrm{COV\text{-}R}_{\tau}
\;=\;
\frac{1}{K_c}\sum_{k=1}^{K_c}
\mathbf{1}\!\left[\max_{g\in\mathcal G}\,\mathrm{TM}(g,\,y_k)\;\ge\;\tau\right]
\quad \uparrow ,
\]
\[
\mathrm{COV\text{-}P}_{\tau}
\;=\;
\frac{1}{K}\sum_{j=1}^{K}
\mathbf{1}\!\left[\max_{y\in\mathcal Y}\,\mathrm{TM}(g_j,\,y)\;\ge\;\tau\right]
\quad \uparrow .
\]

\paragraph{Diversity.}
We quantify ensemble spread as the average pairwise RMSD among generated samples:
\[
\mathrm{diversity}(\mathcal G)
\;=\;
\binom{K}{2}^{\!-1}\!\!
\sum_{1\le i<j\le K}
\mathrm{RMSD}(g_i,\,g_j)
\quad \uparrow .
\]
AMR and COV-R measure recovery of the experimental states, COV-P measures generated-sample fidelity to the experimental ensemble, and pairwise RMSD is used descriptively. We do not treat diversity alone as sufficient evidence of ensemble quality.

\paragraph{Ground-truth conformer distinctness.}
For the bridge ablation, we stratify test clusters by how separated their experimentally resolved conformers are. We measure this separation by average pairwise TM-distance,
\[
D_{\mathrm{GT}}(\mathcal Y)
\;=\;
\binom{K_c}{2}^{\!-1}\!\!
\sum_{1\le i<j\le K_c}
\left(1-\mathrm{TM}(y_i,\,y_j)\right)
\quad \uparrow ,
\]
where larger values indicate more distinct observed conformers.

\paragraph{Conformer-specific recovery stability.}
The recovery-stress analysis uses $K=150$ samples per ablation variant and aggregates results over paired evaluation clusters. We test whether the learned field recovers each supervised conformer from nearby off-path bridge states. Let \(G_1^{(m)}\) denote an experimentally observed conformer of a sequence, and let \(G_t^{(m,\kappa)}\) denote a bridge state around that conformer at time \(t\) with perturbation scale \(\kappa\). Applying the predicted bridge field and the same target-frame recovery used in training gives a recovered conformer \(\hat G_1^{(m,\kappa)}\). We report the target-frame recovery error
\[
E_{\mathrm{frame}}(\kappa)
\;=\;
\frac{1}{|\mathcal M|}
\sum_{m\in\mathcal M}
d_{\mathrm{frame}}\!\left(
\hat G_1^{(m,\kappa)},\,G_1^{(m)}
\right)
\quad \downarrow ,
\]
where \(d_{\mathrm{frame}}\) is the frame-level distance used for the ablation. To test whether this recovery preserves distinct conformational states, we compute the separation retention ratio for conformer pairs,
\[
\mathrm{SepRet}_{a,b}(\kappa)
\;=\;
\frac{
d_{\mathrm{frame}}\!\left(\hat G_1^{(a,\kappa)},\,\hat G_1^{(b,\kappa)}\right)
}{
d_{\mathrm{frame}}\!\left(G_1^{(a)},\,G_1^{(b)}\right)
}.
\]
SepRet values closer to \(1\) indicate that the recovered conformers preserve the ground-truth separation between conformers. Values much below \(1\) indicate collapse toward a shared structure, while values much above \(1\) indicate over-dispersion. We average SepRet over conformer pairs and clusters at each \(\kappa\).

\section{Additional Results}
\label{app:additional-results}

We report additional recovery-stress results at \(t=0.25\) and \(t=0.5\) to complement Figure~\ref{fig:ablation-modes}. At \(t=0.25\), frame-error improvements over Linear Bridge are significant at all tested corruption levels, while improvements over Uniform Stochastic Bridge are significant only at \(\kappa=3\). At \(t=0.5\), frame-error differences are not significant at \(\kappa=0,1\); at \(\kappa=2\), only the comparison with Linear Bridge is significant, and at \(\kappa=3\), both comparisons are significant. At both bridge times, mean SepRet is closer to \(1\) for \our\ across the tested corruption levels, with significant paired differences from both alternatives.

\begin{figure}[!htbp]
\centering
\includegraphics[width=\linewidth]{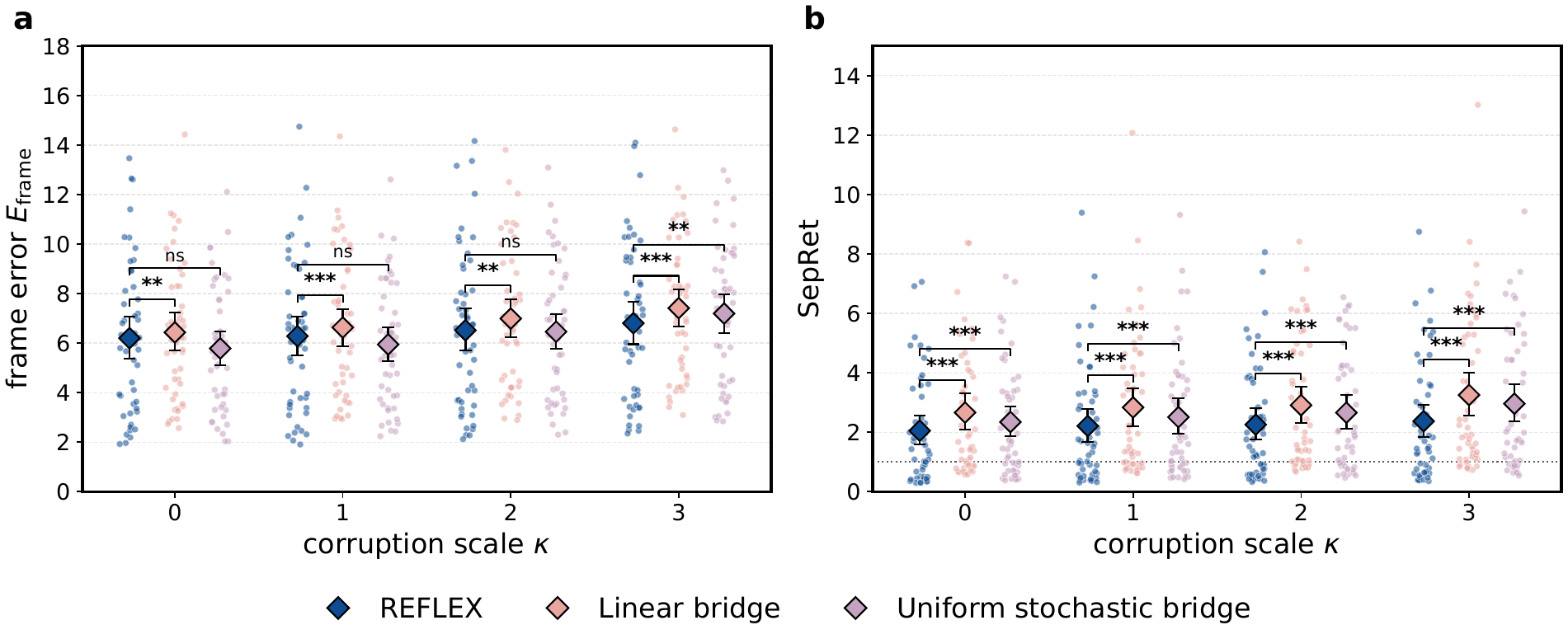}
\caption{\textbf{Additional recovery-stress results at \(t=0.25\).} \textbf{a}: Target-frame recovery error after one-step recovery from perturbed bridge states. \textbf{b}: SepRet under the same perturbation protocol. Stars and \texttt{ns} denote significant and non-significant paired comparisons with \our.}
\label{fig:ablation-modes-t025}
\end{figure}

\begin{figure}[!htbp]
\centering
\includegraphics[width=\linewidth]{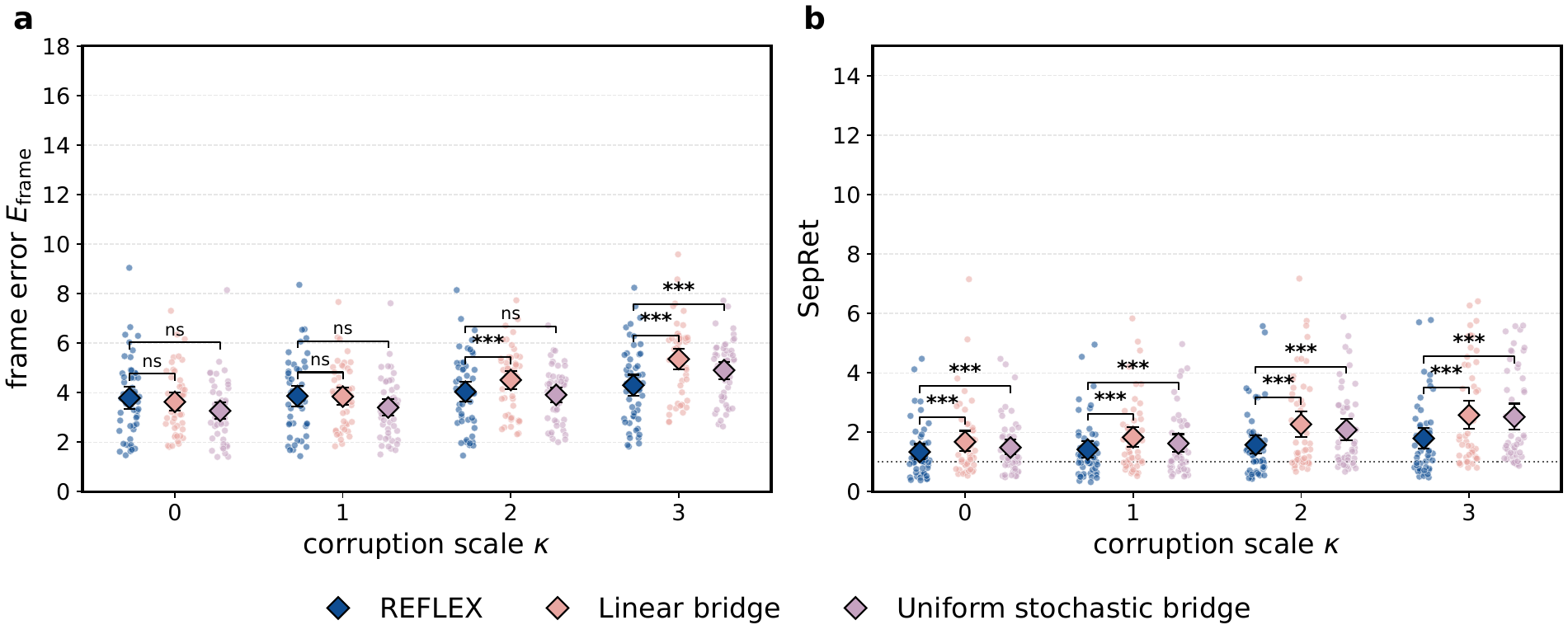}
\caption{\textbf{Additional recovery-stress results at \(t=0.5\).} \textbf{a}: Target-frame recovery error after one-step recovery from perturbed bridge states. \textbf{b}: SepRet under the same perturbation protocol. Stars and \texttt{ns} denote significant and non-significant paired comparisons with \our.}
\label{fig:ablation-modes-t050}
\end{figure}
\FloatBarrier

\section{Statistical Significance Analysis}
\label{app:statistical-analysis}

\begin{figure}[!htbp]
\centering
\includegraphics[width=0.92\linewidth]{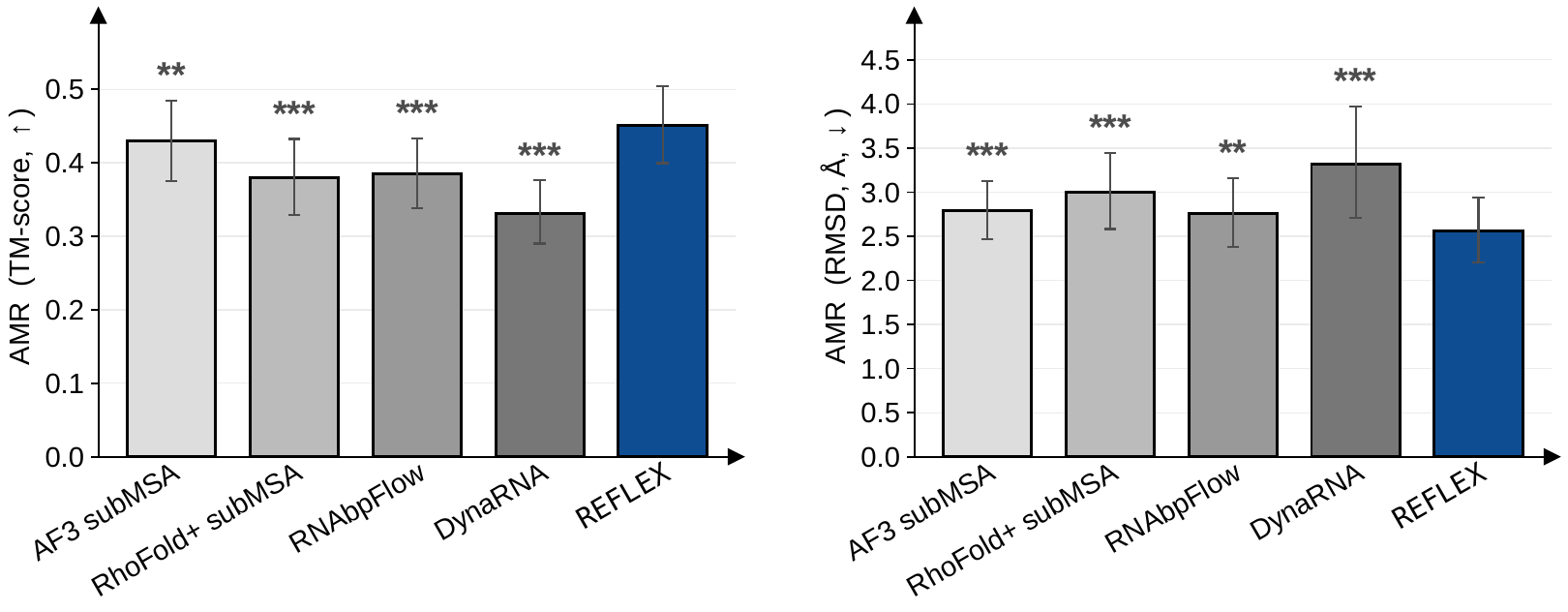}
\caption{\textbf{Statistical analysis of conformer recovery.} Left: best-match TM-score AMR, where higher is better. Right: best-match RMSD AMR, where lower is better. Bars show means with \(95\%\) bootstrap confidence intervals; stars denote Holm-corrected significance from one-sided paired Wilcoxon comparisons with \our.}
\label{fig:amr-statistic}
\end{figure}

\begin{figure}[!htbp]
\centering
\includegraphics[width=\linewidth]{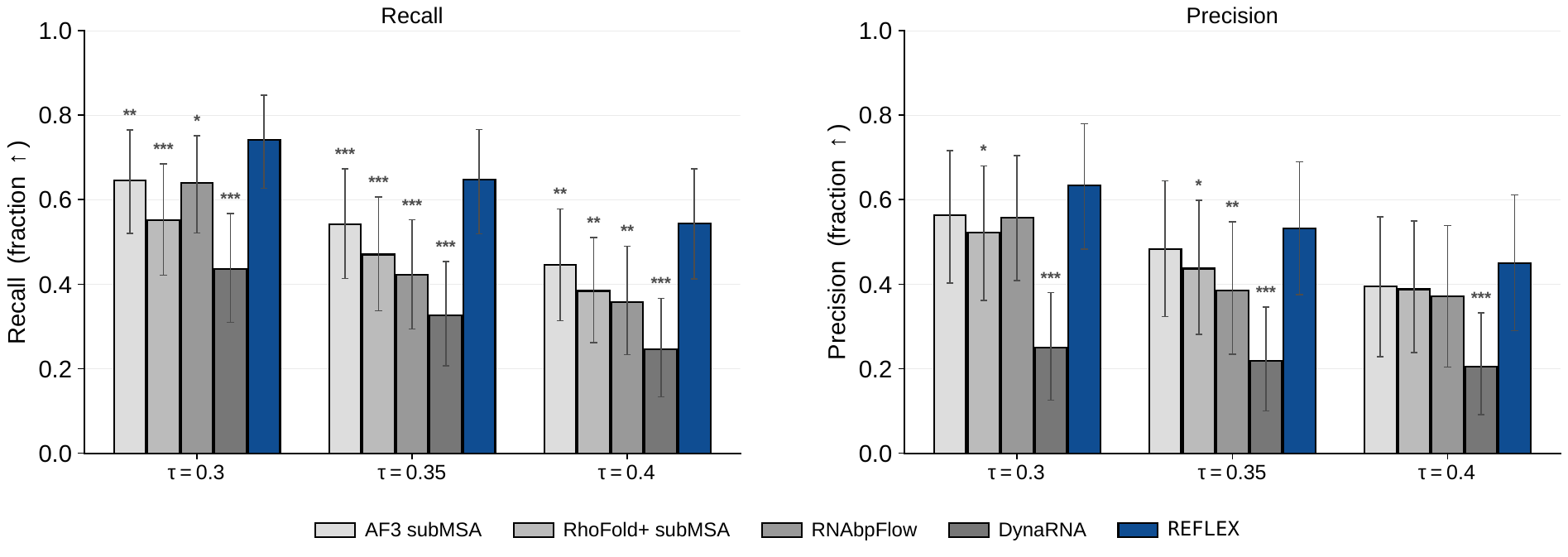}
\caption{\textbf{Statistical analysis for precision--recall coverage.} Left: COV-R at TM-score thresholds \(\tau\in\{0.30,0.35,0.40\}\). Right: COV-P at the same thresholds. Bars show cluster-level means with \(95\%\) bootstrap confidence intervals; stars denote Holm-corrected significance from one-sided paired bootstrap comparisons with \our. Absence of a star indicates that the comparison does not meet the $0.05$ significance threshold.}
\label{fig:cov-statistic}
\end{figure}

For the AMR and coverage analyses, we compute each metric per test cluster and report cluster-level means with \(95\%\) bootstrap confidence intervals from \(10{,}000\) resamples. Significance is assessed against \our\ on the paired cluster intersection. For continuous AMR metrics, we use a one-sided Wilcoxon signed-rank test with the alternative that \our\ is better. For thresholded COV-R and COV-P, we use a one-sided paired bootstrap on the mean cluster-wise difference \(\Delta=\mathrm{REFLEX}-\mathrm{baseline}\). Within each metric, \(p\)-values across baseline comparisons are corrected with Holm's step-down procedure; stars in Figures~\ref{fig:amr-statistic} and~\ref{fig:cov-statistic} denote Holm-corrected significance.

\FloatBarrier

\begin{samepage}
\section{All-atom Reconstruction and Auxiliary Loss}
\label{app:bridge-field-inversion}

We extend the frame-based reconstruction of \textsc{RNA-FrameFlow}~\cite{AnandJMJHMDYHL25} with ten torsions per nucleotide: eight backbone angles, one for $\mathrm{O2}^{\prime}$ placement, and the glycosidic angle $\chi$. Nucleotide-specific reference geometry supplies complete bases, yielding all heavy-atom coordinates:
\begin{equation}
\hat Y = \operatorname{AllAtom}(\hat x_1,\hat R_1,\hat\Phi;s),
\qquad
Y = \operatorname{AllAtom}(x_1,R_1,\Phi;s).
\label{eq:app_backbone_reconstruction}
\end{equation}
\end{samepage}
The auxiliary geometry losses are
\begin{align}
\mathcal L_{\mathrm{atom}}
&=
\frac{1}{|\mathcal A|}
\sum_{a\in\mathcal A}
\|\hat Y_a - Y_a\|_2^2,
\label{eq:app_atom_loss}
\\
\mathcal L_{\mathrm{dist}}
&=
\frac{1}{|\mathcal P|}
\sum_{(a,b)\in\mathcal P}
\Bigl(
\|\hat Y_a-\hat Y_b\|_2 - \|Y_a-Y_b\|_2
\Bigr)^2,
\label{eq:app_dist_loss}
\\
\mathcal L_{\mathrm{tors}}
&=
\frac{1}{|\mathcal T|}
\bigl\|\hat z_{\mathcal T} - z_{\mathcal T}\bigr\|_2^2,
\qquad
z=(\cos\Phi,\sin\Phi),\ \hat z=(\cos\hat\Phi,\sin\hat\Phi).
\label{eq:app_tors_loss}
\end{align}
These terms can be combined as
\begin{equation}
\ell_{\mathrm{target}}
=
\lambda_{\mathrm{atom}}\mathcal L_{\mathrm{atom}}
+
\lambda_{\mathrm{dist}}\mathcal L_{\mathrm{dist}}
+
\lambda_{\mathrm{tors}}\mathcal L_{\mathrm{tors}}.
\label{eq:app_target_loss}
\end{equation}
We use this reconstruction term as \(\mathcal L_{\mathrm{aux}}\) in Eq.~\eqref{eq:lbridge}, while Eq.~\eqref{eq:stochastic_bridge_loss} gives the frame-level stochastic bridge term.

\section{Limitations and Broader Impact}
\label{app:limitations_and_impact}
\our\ uses chain-normalized experimental \(\mathrm{B}\)-factors as weak supervision for relative local mobility during training, so its calibration depends on the availability and quality of experimentally resolved RNA structures. \(\mathrm{B}\)-factors are affected by experimental modality, refinement, resolution, and static disorder, and should not be interpreted as absolute thermodynamic fluctuations. The evaluation is also limited by the scarcity of RNA sequences with multiple experimentally resolved conformers; observed structures may cover only part of the accessible ensemble and do not provide equilibrium populations. Finally, \our\ focuses on sequence-conditioned RNA backbone-frame ensembles, and does not explicitly model ligand-bound switching, protein--RNA complexes, ion effects, or experimentally validated conformational populations.

Future work could explore adapting evolutionary and structure-aware data augmentation tools such as PhyloAug~\cite{cole2025phyloaug} to alleviate the scarcity of multi-conformer RNA training data, with validation of whether augmented examples preserve relevant conformational diversity. Integrating data from complementary modalities, including sequence, structure, and experimental measurements, could further constrain conformational ensembles. Inspired by the separation of common and unique directions in model merging~\cite{wang2026separatingcommonuniquedirections}, a promising direction is to explore how multimodal integration can retain both shared information and modality-specific signals. Future work could also investigate task-aligned adaptation of pretrained RNA representations for conformational ensemble modeling, motivated by findings on the task-dependent transferability of RNA foundation models~\cite{li2026understandingtransfer}.

More broadly, \our\ may help characterize RNA conformational heterogeneity and support hypothesis generation in settings where experimentally resolved ensembles are scarce. However, generated conformers should not be interpreted as experimentally validated structures or functional states. Related concerns motivate knowledge-graph-grounded factuality assessment of medical LLM responses~\cite{zhou2026factchecking} and systematic evaluation of consistency, faithfulness, and biological validity in genomic interpretability~\cite{zhou2026genomicinterpretability}. These perspectives motivate grounding downstream claims in independent evidence and systematically evaluating any mechanistic interpretations derived from generated ensembles. Downstream use may reflect biases in available RNA structures, pretrained representations, and the experimental proxies used for flexibility supervision. Care should therefore be taken when applying the method to biological design, functional interpretation, or therapeutic settings, where generated candidates require empirical validation.


\section{Additional Ablations and Validation}
\label{app:validation}

\paragraph{Matched ablations on ensemble generation.}
We evaluate matched ablations with $K=20$ samples per RNA (Table~\ref{tab:main-ablation}). All variants share frozen AF3 representations, generator backbone, training data, and evaluation protocol. In addition to Linear Bridge and Uniform Stochastic Bridge, we remove flexibility supervision by setting $\lambda_{\mathrm B}=0$ while retaining residue-dependent bridge widths, or remove per-residue calibration by using homoscedastic bridge noise while retaining $\mathcal L_{\mathrm B}$.

\begin{table}[!htbp]
\centering
\caption{\textbf{Matched component ablations.} All variants use $K=20$ samples per RNA. Coverage is macro-averaged across clusters, with the best value in each column in \textbf{bold}. Diversity is average pairwise RMSD in \AA. The two ``w/o'' rows each remove one component from \our.}
\label{tab:main-ablation}
\setlength{\tabcolsep}{3pt}
\begin{tabular}{@{}lccccccc@{}}
\toprule
\multirow{2}{*}{Variant} & \multicolumn{3}{c}{COV-R} & \multicolumn{3}{c}{COV-P} & \multirow{2}{*}{Diversity (\AA)} \\
\cmidrule(lr){2-4}\cmidrule(lr){5-7}
 & @$0.30$ & @$0.35$ & @$0.40$ & @$0.30$ & @$0.35$ & @$0.40$ & \\
\midrule
Linear Bridge & $0.665$ & $0.570$ & $0.449$ & $0.554$ & $0.427$ & $0.383$ & $3.401$ \\
Uniform Stochastic Bridge & $0.687$ & $0.601$ & $0.434$ & $0.606$ & $0.466$ & $0.404$ & $3.088$ \\
w/o flexibility supervision & $\mathbf{0.753}$ & $0.616$ & $0.452$ & $\mathbf{0.647}$ & $0.506$ & $0.410$ & $3.157$ \\
w/o per-residue calibration & $0.717$ & $0.620$ & $0.436$ & $0.620$ & $0.471$ & $0.401$ & $3.252$ \\
\midrule
\our & $0.742$ & $\mathbf{0.647}$ & $\mathbf{0.543}$ & $0.634$ & $\mathbf{0.533}$ & $\mathbf{0.450}$ & $2.900$ \\
\bottomrule
\end{tabular}
\end{table}

\our\ improves all six coverage metrics over Linear Bridge and Uniform Stochastic Bridge while producing lower diversity. Removing per-residue calibration reduces coverage at every threshold. Removing flexibility supervision slightly improves coverage at $\tau=0.30$, but reduces COV-R/COV-P at $\tau=0.40$ from $0.543/0.450$ to $0.452/0.410$. Thus, residue-dependent calibration provides consistent coverage gains, while flexibility supervision mainly benefits recovery at stricter thresholds.

\FloatBarrier

\paragraph{FlexibilityNet prediction.}
We evaluate the $64$-bin supervised prediction head $o_\psi(s)$ on $50$ sequence-disjoint RNA chains with $5{,}857$ annotated nucleotides. Targets are the chain-normalized $\mathrm B$-factor-derived proxy discretized into $64$ bins. We report accuracy within $\pm2$ bins and AUROC for identifying the top $25\%$ of nucleotides within each chain according to this proxy. Metrics are computed per chain and macro-averaged; $95\%$ confidence intervals use $10{,}000$ bootstrap resamples of evaluation chains.

\begin{table}[!htbp]
\centering
\caption{\textbf{FlexibilityNet prediction.} Metrics assess agreement with the relative $\mathrm B$-factor-derived proxy.}
\label{tab:flexibility-validation}
\begin{tabular}{@{}lcc@{}}
\toprule
Metric & Chain-macro result & $95\%$ CI \\
\midrule
Accuracy within $\pm2$ bins & $0.759$ & $[0.691,0.824]$ \\
AUROC for the top $25\%$ by the proxy & $0.681$ & $[0.630,0.731]$ \\
\bottomrule
\end{tabular}
\end{table}

Table~\ref{tab:flexibility-validation} indicates that the supervised head captures informative patterns in the relative proxy. These measurements do not establish agreement with absolute mobility, coordinate RMSF, or thermodynamic fluctuations. At inference, only the latent condition $h_\psi(s)$ is passed to the bridge-field network; the scalar predictions from the supervised head are not consumed by the sampler. The downstream contribution of flexibility grounding is assessed separately by the $\lambda_{\mathrm B}=0$ ablation in Table~\ref{tab:main-ablation}.

\paragraph{Base-pair evaluation protocol.}
\label{app:dssr-validation}
For each experimental conformer, we select the highest-TM-score generated sample, as in AMR-R, without using base-pair information. For \our, these structures include complete nucleobases reconstructed as in \pref{app:bridge-field-inversion}. We annotate reference and matched structures with DSSR~\cite{lu2015dssr}. Base-pair precision, recall, and F1 are computed per match, averaged within each cluster, and then macro-averaged across test clusters.

Results are reported in Table~\ref{tab:dssr-validation} in \pref{sec:exp-structural-recovery}. \texttt{\textsc{DynaRNA}} is excluded because its native output contains only $\mathrm{C4}^{\prime}$ coordinates. DSSR evaluation would require a separate full-atom reconstruction model, introducing dependence on additional post-processing.
\FloatBarrier

\section{Fidelity on RNAs with a Single Deposited Structure}
\label{app:single24}
To assess best-sample fidelity beyond the multi-conformer benchmark, we evaluate 24 RNA chains (71--297 nt) with exactly one deposited experimental structure, using the same temporal cutoff and sequence-identity filtering against the training set. A single deposited conformation does not imply an intrinsically single-state RNA. All methods use the same settings as in the main ensemble benchmark. Experimental structures are used only for evaluation. Predictions are aligned with US-align on C4$'$ atoms (\texttt{-mol RNA}), with TM-score normalized by reference length and RMSD computed over aligned atom pairs.

\begin{table}[!htbp]
\centering
\caption{\textbf{Best-sample fidelity on 24 RNAs with a single deposited structure.} Values are means of per-RNA maximum TM-score and minimum RMSD over US-align-aligned C4$'$ atoms. The best observed values are in \textbf{bold}.}
\label{tab:single24}
\begin{tabular}{@{}lcc@{}}
\toprule
Method & Best TM-score $\uparrow$ & Best RMSD (\AA) $\downarrow$ \\
\midrule
\texttt{\textsc{AF3}} subMSA & $0.435$ & $3.39$ \\
\texttt{\textsc{RhoFold+}} subMSA & $0.429$ & $3.81$ \\
\texttt{\textsc{RNAbpFlow}} & $0.425$ & $3.01$ \\
\our & $\mathbf{0.456}$ & $\mathbf{2.88}$ \\
\bottomrule
\end{tabular}
\end{table}

\our\ achieves the highest observed mean best TM-score ($0.456$) and lowest mean best RMSD ($2.88$\,\AA) among the four methods (Table~\ref{tab:single24}), suggesting that ensemble generation remains compatible with competitive best-sample fidelity on these targets.

\end{document}